\documentclass{article}

\PassOptionsToPackage{numbers, sort&compress}{natbib}
\usepackage{al_insurance}

\usepackage{amsmath}
\usepackage{amssymb}
\usepackage{enumerate}
\usepackage{amsthm}
\usepackage{tikz}

\usepackage[hyphens]{url}

\usepackage{enumitem}

\usepackage{multirow} 
\usepackage{diagbox}

\usepackage{subcaption}

\usepackage{colortbl}
\RequirePackage{etoolbox}
\RequirePackage{xcolor}
\definecolor{cb-black}      {RGB}{  0,   0,   0}
\definecolor{cb-blue-green} {RGB}{  0,  073,  073}
\definecolor{cb-green-sea}  {RGB}{  0, 146, 146}
\definecolor{cb-rose}       {RGB}{255, 109, 182}
\definecolor{cb-salmon-pink}{RGB}{255, 182, 119}
\definecolor{cb-purple}     {RGB}{ 73,   0, 146}
\definecolor{cb-blue}       {RGB}{ 0, 109, 219}
\definecolor{cb-lilac}      {RGB}{182, 109, 255}
\definecolor{cb-blue-sky}   {RGB}{109, 182, 255}
\definecolor{cb-blue-light} {RGB}{182, 219, 255}
\definecolor{cb-burgundy}   {RGB}{146,   0,   0}
\definecolor{cb-brown}      {RGB}{146,  73,   0}
\definecolor{cb-clay}       {RGB}{219, 209,   0}
\definecolor{cb-green-lime} {RGB}{ 36, 255,  36}
\definecolor{cb-yellow}     {RGB}{255, 255, 109}
\definecolor{ao}{rgb}{0.0, 0.5, 0.0}

\usepackage[utf8]{inputenc} 
\usepackage[T1]{fontenc}    
\usepackage{hyperref}       
\usepackage{url}            
\usepackage{booktabs}       
\usepackage{amsfonts}       
\usepackage{nicefrac}       
\usepackage{microtype}      
\usepackage{xcolor}         
\usepackage[english]{babel}
\usepackage{amsmath, amsfonts, amsthm,amssymb,here,dsfont, hyperref}
\usepackage{natbib}
\usepackage{algorithm, algorithmic}

\newcommand{\argmax}[1]{\underset{#1}{\operatorname{arg}\!\operatorname{max}}\;}
\newcommand{\argmin}[1]{\underset{#1}{\operatorname{arg}\!\operatorname{min}}\;}

\newtheorem{theo}{Theorem}[section]
\newtheorem{definition}[theo]{Definition}

\newcommand{\norm}[1]{\left\lVert#1\right\rVert}

\usepackage[textwidth=1.8cm, textsize=tiny]{todonotes}
\title{Fair active learning:\\ Solving the labeling problem in insurance}

\author{%
  Romuald Elie$^1$ \\
  \texttt{romuald.elie@univ-eiffel.fr} \\
  \And
  Caroline Hillairet$^2$\\
  \texttt{caroline.hillairet@ensae.fr} \\
  \And
  François Hu$^{3}$ \\
  \texttt{francois.hu@milliman.com} \\
  \And
  Marc Juillard$^4$ \\
  \texttt{marc.juillard@socgen.com}\\
  \And
  {\normalfont {\small $^1$Universit\'e Gustave Eiffel, $^2$ENSAE-CREST, $^3$Milliman France, $^4$Société Générale Insurance}}
  }

\begin{document}

\maketitle

\begin{abstract}
This paper addresses significant obstacles arising from the widespread use of machine learning models in the insurance industry, with a specific focus on promoting group fairness. The initial challenge is to effectively leverage unlabeled data, reducing the labeling effort while boosting predictive performance. This can be achieved by emphasizing data relevance through active learning techniques. The paper explores various active learning sampling methodologies and evaluates their impact on both synthetic and real insurance datasets. This analysis highlights the difficulty of achieving fair inferences, as machine learning models may replicate biases and discrimination found in the underlying data. To tackle these interconnected challenges, the paper introduces an innovative fair active learning method. The proposed approach samples informative and fair instances, achieving a good balance between model predictive performance and fairness, as confirmed by numerical experiments on insurance datasets.
\end{abstract}


\section{Introduction}
Insurance companies collect and store large quantities of unstructured data on a daily basis, such as insurance agent's free text zones, claims reports, photos of damaged vehicles, emails, customer reviews and more. With the advancements in computing resources and methodologies, Artificial Intelligence (AI) is becoming a driving force for the development and transformation of the insurance industry. Using machine learning (ML) has numerous benefits, including improved risk segmentation, better claims predictions, automated processes, and efficient decision-making. This can enhance risk assessment and operational efficiency while also reducing costs.  
However, managing this large amount of unstructured data involves challenges that are intricately tied to regulatory requirements. Regulatory frameworks, such as the EU rules on gender-neutral pricing in insurance and the General Data Protection Regulation (GDPR) in Europe for data privacy, have a substantial impact on how this data must be handled.
Indeed, the data collected may include information that is either not compliant with GDPR regulations or not aligned with gender-neutral principles, thereby raising valid ethical concerns. Additionally, the potential for ML to propagate unfairness by replicating social and historical biases that exist within the data is also a critical concern~\cite{mehrabi2021survey}.
Currently, the assessment of such data points goes through a costly labeling process done by experts, that is neither fast enough nor scalable. Implementing an accurate, cost-effective, and fair learning system is crucial for the insurance industry to overcome these challenges. Hence, actuaries must seize these new and efficient methodologies to keep and reinforce their expertise of risk evaluation. The objective of this paper is to tackle and resolve several of these challenges by introducing various sampling methods that, when labeled, significantly improve the predictive performance of the model compared to randomly sampling the data. Additionally, the paper explores incorporating a fairness constraint into the sampling strategies to ensure fairer outcomes.

\subsection{Labeling challenge}

Most AI projects in insurance require large amounts of labeled data to train specific algorithms, which in practice are often based on deep learning models and are extremely data-intensive. The labeling step is therefore an important (if not the most important) part of the AI implementation process. 
Today the insurance industry acknowledges the significance of unstructured data accompanied by expert labeling. For instance, insurance companies have identified the potential to utilize telematics data in a multi-class setting to adjust premiums based on driver behavior (also known as pay as/how you drive)~\citep{TSELENTIS2016362, so2021cost}.
High quality labeled data is an important ingredient for learning  to correctly map instances and labels, and the lack of it typically downgrades the model performance, while sometimes introducing biases and ethical problems~\cite{mehrabi2021survey}. 
However, labeling campaigns are usually carried out under time (and therefore volume) and cost constraints. Being able to prioritize the relevant data points to label instead of randomly selecting them is as important in practice as the performance of the learning algorithm that will be trained with these data. These algorithms refer to the field of \textit{active learning} (AL) \cite{settles2009active}. Active learning  is able to sequentially query a human expert (a.k.a. \textit{oracle}) to label new data points with the desired outputs. As we will see in Section~\ref{sec:al}, the key component of AL is to sample informative instances from a pool of unlabeled data.

From theory to practice, the field of AL has received a lot of attention in the recent years, such as  \cite{beluch2018power, sinha2019variational, smith2018less} among others. Furthermore, in the ML literature there exists many AL surveys \cite{settles2009active, fu2013survey, ren2020survey}. However, few works tackle the specific challenges related to actuarial problems, such as the need for fairness and transparency.

\subsection{Active learning in insurance} AL can be especially useful for unstructured data in the insurance industry. Usually, unstructured data (e.g. images of damaged vehicles, telematics data, GDPR compliance, medical reports, accident descriptions, ...) often require more manual efforts and expertise to label compared to structured data. Nevertheless, AL can also be applied to structured or tabular data, leading to numerous practical use cases, and can thus be seen as a tool to enhance traditional actuarial science, including:
\begin{itemize}
    \item \textit{Fraud detection:} Actuaries frequently utilize fraud detection models to detect possible fraud in claims, as fraudulent claims (e.g. vehicle claims) cause significant financial losses for insurance companies. AL can be used to select the most informative examples of (presumably) fraudulent and non-fraudulent claims for expert-labeling, which improves the accuracy of the fraud detection model. For instance, \cite{10.1007/978-3-030-26072-9_11} proposes an AL framework based on model-disagreement (see Section~\ref{subsec:ALdisag}) to combat pension fraud, which can significantly reduce the high-risk population range in detecting fraud cases. 
    Note that fraud detection models should naturally be fair and unbiased in order to prevent policyholders from being unfairly penalized or discriminated against.

\item \textit{Imbalanced dataset:} In scenarios with imbalanced datasets, echoing situations like the fraud detection scenario (another common example being churn detection), where the minority class is rare, AL has proven highly effective. Studies like \cite{ertekin2007learning} provide evidence that methodologies naturally prioritize the sampling of minority classes to improve predictive models.  This approach is particularly significant in insurance, as illustrated by cases like the aforementioned fraudulent case study, where fraudulent activities may not always be easily discernible within the dataset.

    \item \textit{Risk assessment (or customer segmentation):} Insurance companies can use AL to assess risk levels (multi-class task) for different policyholders. By labeling (or relabeling) the most informative and fair examples of policyholders with different risk profiles, the company can develop more accurate models to predict which policyholders are more likely to file a claim and adjust their premiums accordingly. Note that relabeling incorrect or unfair instances can improve the fairness and reliability of the data. For instance, \cite{kumar2010data} applies uncertainty sampling (see Section~\ref{subsec:ALuncer}) to predict and avoid errors in health insurance claims processing, resulting in reduced administrative efforts. Nevertheless, they consider the possibility of using a density-based approach (see Section~\ref{subsec:ALrepre}) for improved outcomes.

    \item \textit{Biased dataset:} In scenarios where labels are influenced by societal or historical biases, such as risk categorization based on demographic factors like race or gender (as seen in the redlining issue \cite{harrington1998race}), it is important to re-label the data impartially. Re-labeling ensures assessment based solely on relevant risk factors such as claims / accidents statistics, rather than historical discriminatory traits.

    \item \textit{Missing values:} Labeling missing values, even in structured data, when information is incomplete could significantly improve the predictive modeling. For example, if a policyholder's driving record has gaps due to errors or omissions, accurately labeling these missing values is essential for risk assessment. Without proper labeling, insurers may make incorrect assumptions or overlook important risk factors, resulting for example in inaccurate pricing and coverage decisions.

\end{itemize}
Overall, active learning can help insurance companies better leverage their data and develop more accurate models for decision making.
The evolving regulatory landscape in insurance adds another layer of complexity, underscoring the necessity for precise and impartial labeling practices to ensure compliance and fairness. Missing labels, alignment with insurance law, and legal aspects in pricing are all important considerations. Addressing these concerns is needed for maintaining the accuracy, fairness, and regulatory compliance of predictive models in insurance, benefiting insurers, policyholders, and regulatory authorities alike.
 While most of the algorithmic fairness literature has focused heavily on "de-biasing" predictive algorithms, it is equally important to support practitioners in selecting and organizing high-quality datasets that prioritize fairness. More detailed discussions on this topic can be found in~\cite{veale2018fairness, holstein2019improving}.

\subsection{Fairness issue}
\label{subsec:BiasAI}

One major drawback of this ever-evolving AI technology is the lack of interpretability of the system. This, together with  the existence of efficient well established actuarial models, has played a role in the slow adoption of AI into the actuarial sciences, in comparison to other fields. In particular, it is hard to evaluate whether the model discriminates or induces unintentional bias, that is not aligned with the ethical standards of the insurance company and may expose it to operational and reputational risks.
Moreover, AI systems learn to make decisions based on labeled data, which can include biased human decisions or reflect social inequities even if the sensitive variables (like gender or age) are removed.
According to~\citep{kearns2019ethical}, ``\textit{machine learning won’t give you anything like gender neutrality ‘for free’ that you did not  explicitly ask for}''.
Hence, one of the biggest challenge in establishing AI governance is the fairness assessment of ML models.

Notably, for EU insurers, the principle of fairness is an obligation and actuaries have often been confronted with this problem \cite{barry2020personalization}.  For instance, it is recognised in the Insurance Distribution Directive (IDD) where it is stated that insurance distributors shall ``\textit{always act honestly, fairly and professionally in accordance with best interests of their customers}'' (Art. 17(1) - IDD). The fairness principle is also stated in Article 5 of GDPR: personal data shall be ``\textit{processed lawfully, fairly and in transparent manner in relation to the data subject}'' (Art. 5(1) - GDPR). 
The concept of bias and algorithmic discrimination is not new, as shown by \cite{abraham1985efficiency, thiery2006fairness, pedreshi2008discrimination}. In particular, \cite{thiery2006fairness} emphasizes that in insurance, insured individuals ``\textit{are classified on the basis of factors that are either immutable such as gender or age, or that are mutable, such as smoking behaviour}''. This topic has been further explored in various research papers on actuarial fairness and pricing discrimination~\citep{guillen2012sexless, landes2015fair, chen2017unisex, frees2023discriminating, charpentier2023mitigating}.
Within these discussions, the debate on fairness includes various ethical viewpoints, with divergent perspectives from regulatory bodies and scholars:
\begin{itemize}
    \item One perspective, which advocates for actuarial fairness, entails assessing insurance premiums' fairness by referencing actuarial rates. According to \cite{landes2015fair}, ``\textit{individuals should pay premiums that reflect the risks they bring to the insurance pool}''. This approach aims to accurately estimate risk by leveraging \textit{all} available data for a more ``\textbf{personalized}'' premium. 

    \item Conversely, the Council of the European Union (2004) advocates for an alternative approach that challenges the notion of actuarial fairness. Instead, it promotes equal treatment, as reflected in the concept of \textit{Demographic Parity} (formally defined later in Section~\ref{subsec:fairdef}), even when risks may differ. This approach emphasizes a form of ``\textbf{solidarity}'', as discussed in \cite{baumann2023fairness, thiery2006fairness}. \cite{thiery2006fairness} mention that ``\textit{this regulation banning discrimination could, in itself, be regarded as a tool to achieve a reasonable balance between the dual goals of actuarial validity and social fairness}''.
\end{itemize}
Following the European Court of Justice's decision on December 21, 2012, EU insurers are no longer allowed to use gender as a factor in determining insurance premiums, resulting companies to potentially reduce personalization levels. While this regulation targets direct discrimination and aims to promote \textit{group fairness}, it does not address indirect discrimination arising from gender-related factors. More specifically in insurance, note that \cite{lindholm2024fair} delineates instances where group fairness like Demographic Parity is aptly applied (e.g., property insurance with gender as a sensitive attribute) and where it may be less pertinent (e.g., commercial accident insurance, which conflicts with actuarial fairness).
These studies underscore the importance for actuaries to remain vigilant regarding fairness issues, particularly given their significant role in establishing insurance prices or any scoring methodologies. For instance, a recent study highlighted in the \textit{Science} journal \cite{obermeyer2019dissecting} revealed ethnic bias in a health score prediction algorithm used in hospitals: ``at the same risk score, Black patients were deemed sicker than white patients''. Rectifying this bias could increase assistance for Black patients ``from 17\% to 46.5\%''. Furthermore, citing from \cite{ito2019supposedly}, we must be aware of ``\textit{How the use of AI runs the risk of recreating the insurance industry's inequities of the previous century}''. In this study, we propose various sampling strategies to strike a balance: enhancing predictive accuracy to improve personalized risk with little labeling cost while ensuring Demographic Parity fairness to serve as a form of ``solidarity''.

Algorithmic fairness can be categorized into \textit{1)} \textit{pre-processing} in enforcing fairness directly in the data, \textit{2)} \textit{in-processing} which enforces fairness in the training step of the learning model and \textit{3)} \textit{post-processing} which reduces unfairness in the model inferences. 
Contrary to the interpretability for which there does not exist yet a clear metric (cf. \cite{molnar2021interpretable, maillart2021toward}), one can formally quantify fairness performance, as detailed in Section~\ref{subsec:fairdef}. 
For actuarial applications, fairness criteria deserve to be taken into account in active learning procedures, as we develop in Section~\ref{sec:alemp} below.

Given the same training-set, it is well-known that enforcing fairness in the model potentially reduces the model predictive performance as shown by both theoretical results~\cite{zafar2019fairness,Agarwal_Beygelzimer_Dubik_Langford_Wallach18, hu2023parametric} and experimental results~\cite{donini2018empirical,barocas2014datas, hu2023sequentially}. Nevertheless, fairness in AI is one of the key challenges for insurers \cite{meyers2018enacting, frezal2020fairness, xin2023antidiscrimination, charpentier2023mitigating} and the fairness evaluation of active learning methods are under-studied. Notably, in their study, \cite{xin2023antidiscrimination} introduced anti-discrimination insurance pricing models, exploring the fairness-accuracy trade-off and emphasizing adverse selection and solidarity. More recently, \cite{charpentier2023mitigating} quantifies and addresses insurance discrimination, offering insights across various insurance use-cases through the use of optimal transport theory. 

Despite the increasing focus on algorithmic fairness in insurance (refer to~\cite{charpentier2023insurance} for additional details or examples in insurance), many proposed methodologies rely on datasets presumed to be complete and informative. However,  the applicability of these studies is limited by the practical challenges posed by the absence of labels, frequently encountered in unstructured datasets such as text, image, or telematics data. This limitation extends even to structured datasets, including those commonly used in insurance premiums, where the lack of labels may occur when information is incomplete or unavailable for assessing risk and determining premium rates for policyholders. In response to this challenge, we introduce a \textit{fair active learning} framework designed to efficiently and fairly label new instances. This approach prioritizes both accuracy and fairness of ML models, resulting in a more equitable and informative dataset. Specifically, we advocate for a classification task applicable to both binary and multi-class scenarios to meet the growing demands in such situations. This is evidenced by the increasing use of unstructured datasets and deep learning methodologies in the insurance industry, as highlighted by~\cite{blier2020machine, hassani2020big}.

Although of high importance, only three recent papers, to our knowledge, investigate the difficulties arising when combining fairness accuracy objectives with active learning algorithms. In addition, none of them focuses on this question from an actuarial perspective. \cite{anahideh2022fair} and \cite{sharaf2022promoting} develop each an algorithm for fair active learning that sample the next data points to be labeled considering a balance between model accuracy and fairness efficiency. On the other hand, \cite{branchaud2021can} studies whether models trained with uncertainty-based (deep) active learning is fairer in their decisions with respect to a sensitive feature (e.g. gender) than those trained with passive learning. It appears that, with neural network architecture, active learning can preemptively mitigate fairness issues.

\subsection{Contributions}
\label{subsec:contrib}

Motivated by the objective of simultaneously improving model predictive performance and fairness, the contributions of the present paper can be summarized as follows:
\begin{itemize}
    \item We review and compare various active 
    learning algorithms used for data labeling and highlights their relevance for actuarial applications.
    
    \item We demonstrate the fairness properties of these methods and introduce a new fair active learning approach. Our method gathers fair and informative instances in the labeling process, and leads to better model performance in comparison to traditional passive learning and better fairness metric in comparison to naive active learning methods.
    
    \item Contrary to other existing methods 
    ~\cite{settles2009active, ren2020survey}, it is worth noting that our proposed approach is model-agnostic and can be applied to any sampling strategy (\textit{query-agnostic}) that uses a probabilistic learning model.

    \item We illustrate the performance of our method on both synthetic and actuarial datasets.
    
\end{itemize}

\paragraph*{Outline of the paper} Section~\ref{sec:notation} formally and independently defines the two problems of labeling and fairness evaluation. For the sake of completeness, we provide in Section~\ref{sec:al} an overview of classical active learning methods, and illustrate their performance on synthetic data in Section~\ref{subsec:illustration}. Section~\ref{sec:FairAL} introduces our fair active learning algorithm for multi-class classification tasks. The performance of  our fair (and traditional) active learning methods on insurance data is discussed in Section~\ref{sec:alemp} .

\section{Problem formulation}
\label{sec:notation}

This section introduces the theoretical framework and mathematical notations. First, we discuss the necessity of (binary and multi-class) classification in insurance. Subsequently, we delve into various measures for assessing the quality of such classification tasks, followed by the presentation of a selected fairness criterion.

\subsection{Multi-class classification task in insurance}

With the recent democratization of ML models in insurance, there has been a notable shift towards incorporating both binary and multi-class classification methodologies in actuarial science~\cite{blier2020machine, gao2019convolutional, liu2014using, richman_2021, so2021cost}. This trend is particularly relevant due to the diverse range of applications seamlessly aligning with the classification paradigm, including claim frequency modeling, risk segmentation, churn prediction, and more. 
 This is particularly  evident in the following cases.
\begin{itemize}
    \item \textit{Insurance pricing:} For instance \cite{liu2014using} predicts claim frequency through multi-class classification, while \cite{gao2019convolutional, so2021cost} utilizes ML models (deep ConvNets for \cite{gao2019convolutional} and Adaboost for \cite{so2021cost}) with telematics data for a multi-class classification framework, identifying high-risk drivers and trips. This last labeling approach provides insurance companies with hazard levels for individual trips, serving as features and risk indicators in insurance pricing.

    \item \textit{Risk classification (leading to insurance coverage and prohibitions):} The insurance industry relies on risk classification to differentiate among risks, and multi-class classification can segment policyholders based on factors like age, driving history, and location, as detailed in \cite{so2021cost}. For example, a policyholder with a clean driving record in a low-crime area may be classified as low-risk, while one with traffic violations in a high-crime area may be labeled high-risk. By leveraging predictive models on historical data, insurers can accurately assign policyholders, enabling coverage decisions (and customized pricing) that improve underwriting accuracy and profitability. However, this approach may raise ethical concerns regarding redlining, where certain demographics may face discriminatory treatment by insurance companies in raising prices and restricting coverage availability, as discussed in \cite{harrington1998race} and \cite{charpentier2023insurance}.

     \item \textit{Fraudulent activity detection:} Detecting fraudulent activities in insurance (or finance) claims is a constant challenge for insurers and multi-class tasks are ever-present \cite{kim2016detecting, anowar2019multi}. Multi-class classification models can for instance help distinguish between legitimate claims, suspicious claims requiring further investigation, and clearly fraudulent claims. By analyzing various attributes of claims data, such as claimant behavior, claim details, and historical patterns, these models can identify different classes of fraudulent activity and assist insurers in mitigating financial losses due to fraud.

    \item \textit{Business optimization:} 
Aside from classical actuarial science methodologies, which focus on modeling risk in insurance and finance, multi-class classification is essential in other modeling areas such as claims categorization and customer segmentation. In claims categorization, insurance companies face diverse claims ranging from auto accidents to health emergencies. Multi-class classification streamlines this process by categorizing claims automatically, ensuring efficient handling of each type. Similarly, in customer (or risk) segmentation, understanding behavior and preferences is crucial for tailored product offerings. Multi-class classification segments customers into groups like young families or retirees, allowing insurers to customize marketing campaigns effectively.
    


\end{itemize}

Finally, the most evident application of multi-class modeling is in dealing with unstructured data~\cite{blier2020machine, richman_2021}, such as images of damaged vehicles or the analysis of texts from GDPR compliance or accident descriptions --- scenarios increasingly prevalent in the insurance industry.

\subsection{Theoretical and empirical misclassification risk}
Let us denote $\mathcal{X}$ the space of instances and $\mathcal{Y}$ the space of labels (or classes). We also consider the adequate probability measures for these spaces: $\mathcal{P}$ the distribution over $\mathcal{X}\times \mathcal{Y}$ and $\mathcal{P}_\mathcal{X}$ the marginal distribution of $\mathcal{P}$ over $\mathcal{X}$. We denote $\mathcal{H}: \mathcal{X}\to \mathcal{Y}$ the space of hypotheses (also called the set of predictors). For a given labeled instance $(x, y)\in\mathcal{X}\times \mathcal{Y}$ and a given predictor $h\in\mathcal{H}$, $\hat y = h(x)$ is the prediction of the label of $x$.

A prediction is evaluated by a loss function denoted $l: \mathcal{Y} \times \mathcal{Y} \to [0, +\infty[$. For instance, a classical way for a classification problem (i.e. a task restricted to a finite set of classes $\cal{Y}$) is to choose the \emph{misclassification loss} defined by $l_{0-1}(y, y') = \mathds{1}(y \neq y')$ and the \emph{square loss} for regression tasks (i.e. our task is to predict a continuous output) defined by $l_{2}(y,y') = (y - y')^2$. 

Note that the loss function evaluates the performance of a predictor on a single observation. To evaluate the predictors on a set $\mathcal{X}$ we define the following (theoretical) risk function: for any loss function $l$, and denoting $X$ (resp.  $Y$) the instance (resp. label) random variable, we have
\begin{equation}\label{eq:ThRisk}
    R(h):= \mathbb{E}[l({ h(X), Y})] = \int_{\mathcal{X}\times\mathcal{Y}} l(h(x), y)\mathcal{P}(x,y)dxdy.
\end{equation}

In practice, the distribution $\mathcal{P}$ is often unknown, therefore, it is analytically intractable. A (natural) estimator of this risk is its \emph{empirical risk}: If we denote $(x_i, y_i)_{i=1}^{N}\sim \mathcal{P}^N$ the observations then
\begin{equation}
\label{eq:EmpRisk}
\hat R(h):= \frac{1}{N}\sum\limits_{i=1}^{N} l(h(x_i), y_i).
\end{equation}

Let us consider the following misclassification risk:

\emph{Theoretical risk}: it represents the probability that the predictor predicts a different answer than the oracle
\begin{equation*}
R(h) =  \mathbb{E}[\mathds{1}(h(X) \neq Y)] = \mathbb{P}(h(X) \neq Y).
\end{equation*}

\emph{Empirical risk}: it represents the average of times the predictor
misclassifies on the data
\begin{equation}
\label{eq:defR}
\hat R(h) = \frac{1}{N}\sum\limits_{i=1}^{N}
\mathds{1}(h(x_i) \neq y_i)).
\end{equation}

\subsection{Active sampling and model predictive performance}
Let $\mathcal{D}^{(train)} = \{ (x_i^{(train)}, y_i^{(train)})\}_{i=1}^L$ be the training-set and $\mathcal{D}^{(test)} = \{(x_i^{(test)}, y_i^{(test)})\}_{i=1}^T$ the test-set where $(x_i, y_i)$ are drawn i.i.d. according to the distribution $\mathcal{P}$. We assume that a large pool of unlabeled  dataset denoted $\mathcal{D}_\mathcal{X}^{(pool)} = \{x_1^{(pool)}, \dots, x_U^{(pool)}\}$ (called pool-set) is available. The statistical learning process consists in the following three steps:

\begin{enumerate}
    \item \textbf{Querying (or labeling) step.} This step involves labeling the raw data to create labeled data for both training and test-sets. Two types of labeling are considered: \textit{passive labeling}, where data is randomly queried, and \textit{active labeling}, where data is queried based on an importance criterion. Passive labeling generates an i.i.d. database, while active labeling does not guarantee independence. The hold-out set (test data) should be generated using passive labeling to ensure that it replicates $\cal{X}\times \cal{Y}$ empirically.
   
    \item \textbf{Training step.}  
   As computing the minimizer $h^*$ of the theoretical risk in Equation~\eqref{eq:ThRisk} is impractical, we use \textit{Empirical Risk Minimization} (ERM) to minimize its empirical form, which is given by Equation~\eqref{eq:EmpRisk}. Given the training-set $\mathcal{D}^{(train)}$, our aim is to find
    
    \begin{equation*}
        \hat h = \argmin{h\in\cal{H}} \hat{R}_{train}(h)
    \end{equation*}
    where
    \begin{equation*}
    \hat R_{train}(h):= \frac{1}{L}\sum\limits_{i=1}^{L} l\left( h(x_i^{(train)}), y_i^{(train)} \right).
    \end{equation*}
    After this step we expect that minimizing the empirical risk is approximately equivalent to minimizing the true risk:
    $$
    \hat{R}_{train}(\hat h) \approx {R}(h^*).
    $$
    
    \item \textbf{Testing step.} This step evaluates the performance of estimator $\hat h$ on the hold-out set $\mathcal{D}^{(test)}$ and detect overfitting or underfitting on the training-set. $\hat h$ must satisfy the following condition:
    \begin{equation*}
        \hat{R}_{train}(\hat h) \approx  \hat{R}_{test}(\hat h)
    \end{equation*}
    where
    \begin{equation*}
        \hat R_{test}(h):= \frac{1}{T}\sum\limits_{i=1}^{T} l\left( h(x_i^{(test)}), y_i^{(test)} \right).
    \end{equation*}
    \end{enumerate}

Given a classifier $h\in\mathcal{H}$, when fairness is required, two important aspects of the multi-class classifier need to be controlled: its misclassification risk $R$  defined in Equation~\eqref{eq:defR} and its unfairness that will be discussed in the next section.

\subsection{Model unfairness}
\label{subsec:fairdef}


As mentioned in Section~\ref{subsec:BiasAI}, reducing the unfairness of models is one of the biggest challenge in insurance. Let us first rigorously define the unfairness metric, that we are considering here.

Let $\mathcal{S} = \{ -1 , +1 \}$ be the space of a sensitive feature (e.g. gender). We also consider $\cal{X}_{-S}:= \cal{X}\setminus\cal{S}$ the space of instances excluding the space of the sensitive feature, $K$ the total number of classes and $Y \in [K]:=\{1, \ldots, K\}$ the true response of the task. 
Several notions of fairness have been considered~\cite{zafar2019fairness,barocas2014datas}. All of them impose some independence condition between the sensitive feature and the prediction. 
This independence can be desired on some values of the label space (as it is the case for \emph{Equality of odds} or \emph{Equal opportunity}~\citep{hardt2016equality}). In this paper, we focus on the \emph{Demographic Parity}~\citep{calders2009building} that requires the independence between the sensitive feature and the prediction, while not relying on labels. 
Following~\citep{denis2024fairness}, the Demographic Parity in multi-class tasks requires to have non-discrimination group in all predicted outcomes.

\begin{definition}[Demographic Parity]{}
 \label{def:unfairnessDP}
In Demographic Parity (a.k.a. Statistical Parity), we say that a classifier $h\in\mathcal{H}$ is fair (or exactly-fair) with respect to the distribution $\mathbb{P}$ on $\mathcal{X}_{-S} \times \mathcal{S} \times [K]$ if

\begin{equation*}
\mathbb{P}\left(h(X, S) = k|S=1\right) = \mathbb{P}\left(h(X, S) = k|S=-1\right) \,, \quad \forall k\in[K]  \enspace.
\end{equation*}

This definition states that privileged and non-privileged group should have equal likelihood. Note that in an approximate version one can define the notion of $\epsilon$-fairness as for a given $\epsilon > 0$,
\begin{equation*}
|\mathbb{P}\left(h(X, S) = k|S=1\right) - \mathbb{P}\left(h(X, S) = k|S=-1\right) | \leq \epsilon \,, \quad \forall k\in[K] .
\end{equation*}

\end{definition}

In rendering the model output and the sensitive feature independent, Demographic Parity has a recognized interest in various applications, such as for risk segmentation without gender attributes~\cite{guillen2012sexless, chen2017unisex} or for crime prediction without ethnic discrimination~\citep{Hajian_2011_discrimation,KamiranZC13RemoveIllegal,barocas2014datas}.

\paragraph*{Evaluation in Demographic Parity} From the above expression, the unfairness of the model is naturally evaluated by
\begin{equation}
 \label{eq:unfairnessDP}
\mathcal{U}(h):= \max\limits_{k\in[K]} \left|\mathbb{P}\left(h(X, S) = k | S=1 \right) - \mathbb{P}\left(h(X, S) = k| S=-1\right) \right|.
\end{equation}
A classifier $h$ is more fair as $\mathcal{U}(h)$ becomes small. This quantity being intractable, its empirical form will be used in practice in  Section~\ref{subsec:metrics}.



\section{Active learning methods}
\label{sec:al}
 
For the present study let us introduce the active learning setting before studying its sampling methods.

\subsection{Definitions and framework}

Active learning iteratively queries an oracle to label instances in a pool-set, providing the most informative instances for learning a hypothesis $h\in \cal{H}$. The model is trained and updated after each query. Access to unlabeled data can be obtained through various methods including:

\begin{enumerate}
    \item the \emph{offline scenario}, where raw data is directly accessible in large quantities (e.g. free text fields used by teleconsultants, customer reviews, ...), and it involves \textit{pool-based sampling}. The goal is to query the most informative instances from the unlabeled data $\mathcal{D}_\mathcal{X}^{(pool)}$ using a trained model $h\in\mathcal{H}$ and an importance score $I(\cdot, h)$. \textit{Batch-mode sampling} refers to querying multiple instances per iteration, with a natural approach being to query the top instances based on the importance score.

    \item 
    the \emph{online scenario} involves collecting data one by one (e.g. incoming emails), using a query strategy called \textit{stream-based sampling} (or \textit{online sampling}). In this scenario, the active learner decides whether to query each new instance $x^{(stream)}$ or not, obtaining the label $y^{(stream)}$ for accepted queries via the oracle. An importance score function can be used to decide which instances to query, often by setting a threshold for the score.
\end{enumerate}

The objective of active learning in both scenarios is to retain the optimal set of queries that maximizes model performance (i.e., minimizing the empirical risk). In contrast, the traditional model is trained on a dataset randomly queried from the pool-set. This process is called \textit{passive learning} (PL). 

In active learning, the objective is to build a more accurate model than the passive learning approach. This means that the model trained through active learning, denoted as $\hat h_a$, is expected to have a lower test error than the model trained through passive learning, denoted as $\hat h_p$, with the same training-set size: $\hat R_{test}(\hat h_{a}) < \hat R_{test}(\hat h_{p})$. Active learning has several theoretical guarantees under some conditions (see the \textit{Massart noise} in~\citep{balcan2009agnostic}) and can greatly outperform passive learning~\cite{hanneke2014theory}.

\paragraph{About the training step} Let $h_t\in\cal{H}$ be the current trained model at step $t$. We assume that $(x^{*}, y^{*})$ is the current queried data. During the  $t+1$-th training step of the active learner, we can either 1) construct $h_{t+1}$ in re-training $h_{t}$ on the whole labeled set including $(x^{*}, y^{*})$ (called \textit{batch learning} approach) or 2) construct $h_{t+1}$ in updating the weight of $h_{t}$ based only on $(x^{*}, y^{*})$ (called \textit{online learning} approach, see~\citep{shalev2011online} for more details). 

\paragraph{About the learning task} The sampling strategy used in active learning depends on the learning task. For classification problems, the sampling strategies differ from those used in regression due to the nature of the model responses. Probabilistic classification models give interpretable responses in terms of uncertainty, leading to some natural heuristic choices for the importance score function.

In practice, although some tabular data may require labeling (see Section~\ref{sec:alemp}), active learning is particularly useful in labeling unstructured data such as texts, images, and sounds, which can be difficult, time-consuming, and costly to label. Labeling unstructured data requires the intervention of human experts (hence the \textit{cost}) that may require some expertise (hence the \textit{difficulty}) to label the data one by one (hence the \textit{time}).

\begin{figure}[H]
\begin{center}
\includegraphics[scale=0.54]{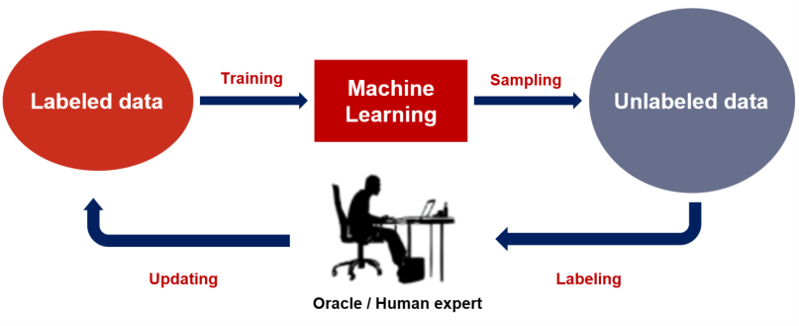}
\caption{Active learning in an offline scenario.}
\label{fig:AL}
\end{center}
\end{figure}

\paragraph*{Settings} Motivated by the above practical insights we study the active learning for classification tasks. Furthermore, we are in an offline scenario: we assume that unlabeled data is inexpensive and abundant but labeling them is difficult, expensive and time-consuming. Let $I:\mathcal{X}\times \mathcal{H} \to \mathbb{R}$ be an importance score function that gives a score to an instance according to its \textit{informativeness}. In the next sections we will introduce the various classical approaches in active learning to define the importance score $I$ before adding a fairness component. Some specific approaches are studied on synthetic and real datasets. We present in Algorithm~\ref{alg:activelearning} the \textit{model-agnostic} active learning process (see Figure~\ref{fig:AL}). We call model-agnostic an approach one can use for any given machine learning model.

\begin{algorithm}[H]
   \caption{Outline of an (model-agnostic) active learner process in an offline scenario}
   \label{alg:activelearning}
\begin{algorithmic}
   \STATE {\bfseries Input:} $h$ a base estimator, $\mathcal{D}^{(train)}$ the initial training-set and $\mathcal{D}_\mathcal{X}^{(pool)}$ the initial pool-set.\medskip
   
   \STATE {\bf \quad Step 0.} Fit $h$ on the training-set $\mathcal{D}^{(train)}$.\medskip
   
   \STATE {\bf \quad Step 1.} The active learner queries the instance $x^{*} \in \mathcal{D}_\mathcal{X}^{(pool)}$ that maximizes the importance score function $I(x, h)$ i.e.
   
    \begin{equation*}
        x^{*} = \argmax{x \in \mathcal{D}_\mathcal{X}^{(pool)}} \left\{ I(x, h) \right\}
    \end{equation*}
   
   \STATE {\bf \quad Step 2.} Update the training-set and the pool-set: if we denote $y^{*}$ its label then 
   
   \begin{eqnarray*}
       \mathcal{D}^{(train)} & = & \mathcal{D}^{(train)} \cup \{(x^{*}, y^{*})\}\\
       \mathcal{D}_\mathcal{X}^{(pool)} & = & \mathcal{D}_\mathcal{X}^{(pool)} - \{x^{*}\}
   \end{eqnarray*}
   
   \STATE {\bf \quad Step 3.} As long as we do not reach a stopping criterion (e.g. exhaustion of the labeling budget or convergence of the performance), we repeat this process (\textbf{return to step 0}).\medskip
   \STATE {\bfseries Output:} the final estimator $h$; the dataset $\mathcal{D}^{(train)}$.
\end{algorithmic}
\end{algorithm}

\subsection{Sampling strategies}
\label{subsec:sampling}

Given the current predictor $h_t\in\mathcal{H}$ ($t$ being the $t-th$ update of the model) and an instance $x\in\mathcal{X}$, we denote $p_{t, x}(y):=\mathbb{P}(h_t(x) = y|x)$ the probabilistic response of classifying $x$ as $y$ according to $h_t$. We consider $\mathcal{D}^{(train)}_{t}$ the data labeled at iteration $t$.
The goal is to make the updated model $h_{t+1}$ perform better than the non-updated model $h_t$ or better than passive learning.

\subsubsection{Sampling based on uncertainty}
\label{subsec:ALuncer}

This approach considers that the most informative instances are the ones located in the region of uncertainty of the learning model. The idea is therefore to query the sample whose prediction is the most uncertain according to the estimator. This approach tends to avoid querying redundant instances since the model learned on an uncertain data point will probably be more certain. We call this approach \textit{uncertainty sampling}.

\paragraph{Binary classification} In a binary task $\mathcal{Y} = \{0, 1\}$, a natural method is to sample the least confident instances (LC) for the learning model: for a given $y\in\mathcal{Y}$,
\begin{equation*}
\hat x_{\rm{LC}} = \argmin{x\in\mathcal{D}_\mathcal{X}^{(pool)}}\left\{\left|\  p_{t, x}(y) - \frac{1}{2} \ \right| \right\}
\end{equation*}
where the \textit{un}importance criterion is defined by $I_{\rm{LC}}(x, h_t) = \left|\  p_{t, x}(y) - \frac{1}{2} \ \right|$. The queried instance $\hat x_{\rm{LC}}$ is in the region of uncertainty of $h_t$. 

\paragraph{Multi-class classification} In a multi-class task $\mathcal{Y} = [K]:= \{1, \dots, K\}$ with $K > 2$ the number of classes, a naive approach would be to generalize the approach presented in the binary case by querying the least confidence instances:
\begin{equation*}
\hat x_{LC} = \argmin{x\in\mathcal{D}_\mathcal{X}^{(pool)}}\left\{p_{t, x}(y^{(K)}) \right\}
\end{equation*}
where we denote $y^{(k)}$ the argument of the $k$-th highest value of $p_{t, x}$. It follows $p_{t, x}(y^{(1)}) \leq p_{t, x}(y^{(2)}) \leq \dots \leq p_{t, x}(y^{(K)})$ with $y^{(K)}$ being the highest probable class for $x$ under the model $h_t$,
\begin{equation*}
    y^{(K)} = \argmax{y\in [K]} \ p_{t, x}(y).
\end{equation*}
Thus $\hat x_{LC}$ is the queried instance with the smallest probability of the most probable class. Note that this approach focus only on $y^{(K)}$ and ignores the distribution of the other classes $\{y^{(k)}, \ k \neq K\}$ and therefore, it raises the following issue: for a given instance $x$ if the margin between $p_{t, x}(y^{(K)})$ and $p_{t, x}(y^{(K-1)})$ is small then $x$ can be considered as an \textit{uncertain} data point \textit{undetected} by the $LC$ criteria. In contrast, a given instance can be considered as a \textit{certain} data point if there is a large margin between class probabilities. The Shannon entropy (defined in \citep{shannon1948mathematical}) overcomes this problem.

The Shannon entropy is often used in statistical learning as a measure of uncertainty. 
In this case, the Shannon entropy assigns an uncertainty score to a given instance $x$ according to the distribution of $\bold{p_{t,x}} = \left(p_{t,x}(1), p_{t,x}(2), \cdots, p_{t,x}(K) \right)$. Each event $\{Y=y\}$ occurs with probability $p_{t,x}(y)$. This measure classifies instances according to whether the posterior probability distribution of the labels is uniform. More precisely  uniform distribution makes it hard for the model to decide on the right class, while a non-uniform distribution leads to a more reliable prediction. 
The uncertainty sampling with the Shannon entropy samples
\begin{equation*}
    \hat x_H = \argmax{x\in\mathcal{D}_\mathcal{X}^{(pool)}}\left\{ I_H(x, h_t) \right\}
\end{equation*}
with $I_H(x, h_t) = -\sum\limits_{k=1}^{K}p_{t, x}(k)\log p_{t, x}(k)$. We use the convention $0\log 0 = 0$.

\paragraph*{Remark}
We denote $I_Y(.) = -\log p_{t,x}(.)$ the information (or \textit{self-information}) that the random event $\{Y=.\}$ contains. Thus defined, the more uncertain an event is, the more informative it is. 
We note that the entropy can be written as $I_H(x, h_t) = \mathbb{E}[-\log p_{t,x}(Y)] = \mathbb{E}[I_Y]$.
The Shannon entropy of a random variable $Y$ corresponds to the expectation of the information contained in the variable $Y$.

Although these uncertainty-based methods can strengthen the learning model, especially on regions of uncertainty, they remain difficult to use when the model is not reliable enough for prediction. Indeed, the current training-set may not contain sufficient information for the model: we call this the \textit{cold-start problem} 
\citep{yuan-etal-2020-cold, jin2022cold}. Moreover, this sampling method works only for probabilistic model and therefore, is not adequate for deterministic or non-probabilistic models (e.g. Support Vector Machine a.k.a. SVM). Alternative methods exist to overcome (at least partially) these problems such as disagreement-based sampling.

\subsubsection{Sampling based on disagreement}
\label{subsec:ALdisag}

For an active learning process the choice of the initial labeled data is crucial so as to get an accurate uncertainty score. If the early dataset is insufficiently representative of $\cal{X}$ then too many plausible model parameters can be suggested for fitting such dataset leading therefore to a \textit{model robustness} issue. We call this phenomenon a \textit{high epistemic uncertainty}~\citep{hullermeier2021aleatoric, nguyen2019epistemic}. By contrast a \textit{low epistemic uncertainty} indicates a robust model. Usually epistemic uncertainty arises in areas where there are not enough instances for training.

Instead of relying on the uncertainty measure based on a single model, disagreement-based sampling proposes a more robust method by combining the result of several learning models all different from each other (so-called \textit{ensembling} methods~\citep{dietterich2000ensemble, denuit2019effective}). The idea described in \cite{seung1992query} is to rely on a set of models to "vote" the informativeness of each instance. This set of models is named a \textit{Committee}. For a Committee an informative instance is characterized by the highest voting disagreement among the models. For a given instance, a model's vote can be characterized either by its label prediction (a.k.a. \textit{hard vote}) or by its label posterior probability (a.k.a. \textit{soft vote}). This approach is called \textit{Query-By-Committee} (a.k.a. \textit{QBC}). QBC requires two main components, namely the construction of the \textbf{Committee} and the definition of the \textbf{disagreement measure} answering respectively the following questions: 1) how can we define a set of models close enough to sample the adapted regions of uncertainty but different enough to ensure the informativeness of the queries? 2) how to measure the degree of disagreement of the Committee?

\paragraph{Committee} Consider a committee of models $h_{t}^{\rm{committee}} = \{h_t^1, \dots, h_t^C\}$ where each $h_t^i\in \mathcal{H}$ is trained on the current training-set $\mathcal{D}_t^{(train)}$. QBC is a query strategy based on the maximum disagreement of $h_{t}^{\rm{committee}}$ constructed by randomized copies of a learned model. At each iteration, the algorithm generates a new committee of classifiers based on the updated training-set.

Initially in his simulations \cite{seung1992query} has implemented a Gibbs training to learn two perceptrons: each perceptron is consistent with the training-set $\mathcal{D}_t^{(train)}$ with slightly different parameters due to the randomness of the Gibbs algorithm. These generated models query the instance where their predictions are the most dispersed. However Gibbs algorithm has a high computational complexity and is hard to implement for more complex learning models. We overcome this problem by using other approaches to generate $h_t^{\rm{committee}}$. 
For instance, Ensemble methods like Bagging or Boosting are often used to construct the committee in discriminative (e.g., logistic regression) or non-probabilistic (e.g., SVM) models. These query strategies were first proposed by \cite{abe1998query}.

$\bullet$ \textbf{Query-by-bagging} (QBag). QBag involves bootstrapping the current training-set into $C$ sets of the same size, denoted by $\mathcal{D}_t^{(train), 1}, \dots, \mathcal{D}_t^{(train), C}$. A committee of models $h_t^{\rm{committee}} = \{h_t^1, \dots ,h_t^C\}$ is then constructed, with each member $h_t^{i}$ corresponding to the base model $h$ trained on $\mathcal{D}_t^{(train), i}$.

$\bullet$ \textbf{Query-by-boosting} (QBoost). 
QBoost involves building a committee of models $h_t^{\rm{committee}} = {h_t^1, \dots, h_t^C}$ using \textit{Adaboost} \cite{FREUND1997119} on the base model $h$, where $(h^1, \dots, h^C)$ is a sequence of copies of $h$ trained on iteratively modified datasets. The approach selects an instance for which the weighted vote obtained by boosting is the most dispersed.

\paragraph{Disagreement measure} Given a Committee $h_{t}^{\rm{committee}} = \{h_t^1, \dots, h_t^C\}$
, there exists various disagreement measures for evaluating the dispersion of votes for multi-class tasks, such as:
\begin{itemize}
    \item \textit{Vote by entropy} \cite{Dagan95committee-basedsampling}, is an entropy-based method combined with a \textit{hard} committee vote:
    \begin{equation*}
        \hat x_{VE} = \argmax{x\in\mathcal{D}_\mathcal{X}^{(pool)}}\left\{ -\sum\limits_{k=1}^{K} \frac{v^{\rm{committee}}_{t, x}(k)}{C} \log \frac{v^{\rm{committee}}_{t, x}(k)}{C} \right\}
    \end{equation*}
    where $v^{\rm{committee}}_{t, x}(k):= \sum\limits_{c=1}^{C} \mathds{1} \{h_t^{c}(x) = k\}$ is the number of \textit{hard} votes of the committee for the label $k$ given the instance $x$.
    
    \item Mean \textit{Kullback-Leibler (KL) divergence} \cite{mccallumzy1998employing}, is a method based on the KL divergence combined with a \textit{weak} committee vote:
    \begin{equation*}
        \hat x_{KL} = \argmax{x\in\mathcal{D}_\mathcal{X}^{(pool)}}\left\{ \frac{1}{C}\sum\limits_{c=1}^{C} D(\ p^{c}_{t,x} \ ||\  p^{committee}_{t, x}\ ) \right\}
    \end{equation*}
    where for all $c \in [C]$, $D(\ p^{c}_{t,x} \ ||\  p^{committee}_{t, x}\ )$ is the KL divergence defined by
    \begin{equation*}
        D(\ p^{c}_{t,x} \ ||\  p^{committee}_{t, x}\ ) = \sum\limits_{k=1}^{K} p^{c}_{t,x}(k) \log \left\{\frac{p^{c}_{t,x}(k)}{ p^{committee}_{t, x}(k) }\right\}
    \end{equation*}
    with $p^{c}_{t,x}(k) = \mathbb{P}(h^{c}_t(x) = k|x)$  and $p^{committee}_{t,x}(k) = \frac{1}{C}\sum\limits_{c'=1}^{C}p^{c'}_{t,x}(k)$ an averaged (over all committee members) probability that $k$ is the correct class.
\end{itemize}

In Sections~\ref{subsec:ALuncer} and \ref{subsec:ALdisag}, we covered techniques for querying uncertain instances in classification, characterized by classifier uncertainty or disagreement among a model committee. These methods enhance local prediction quality but may overlook the influence of queried instances on other parts of the input space. The subsequent sampling strategies advise querying instances with the most significant impact on learning models.

\subsubsection{Sampling based on model change}
\label{subsec:ALimpac}

In the current literature~\citep{Roy2001TowardOA, SettlesCraven08, cai2013maximizing, freytag2014selecting, yoo2019learning}, the idea of these approaches is to choose the instance that gives the most change (or impact) on our learning model if we know its label. If we know its label, a candidate instance can impact the model mainly in two manners: 1) \textit{impact on the model parameters} and 2) \textit{impact on the model performance}.

\paragraph*{Impact on the model parameters}

The idea is to query instances that can change the learning model $h_t\in\mathcal{H}$ as much as possible. This can be done by evaluating the change of the model parameters between the updated model $h_{t+1}$ and the current model $h_t$. Intuitively, if an instance is able to modify considerably the parameters of a model, then this instance contains information on the underlying distribution $\mathcal{X}$ which is not (or rarely) found in the training-set. In the following, we call this set of strategies the \textit{Expected Model Change} (EMC). An example of a change measure is the \textit{Expected Gradient Length} (EGL) given in~\citep{SettlesCraven08}.
    
The EGL strategy applies to all learning models that require the computation of the gradient of a loss function during training (e.g. training via gradient descent). We denote $l_{t}$ a loss function with respect to the model $h_t\in\mathcal{H}$ and $\nabla l_t$ its gradient. The score of importance is measured by the Euclidean norm of the training gradient: 
a \textit{small impact} results in a norm $\norm{\nabla l_t(\mathcal{D}^{(train)}_{t+1})} - \norm{\nabla l_t(\mathcal{D}^{(train)}_{t})} \approx 0 $ while a \textit{large impact} results in a large margin between these two norms. As we do not know the label in advance, the instance to query $x$ corresponds (on average over all possible labels) to a larger gradient size $\mathbb{E}\left[ \norm{ \nabla l_{t}\left(\mathcal{D}_t^{(train)} \cup ( x, Y) \right) } \right]$ with $Y \sim p_{t,x}$ a random variable on $\mathcal{Y}$. That is
\begin{equation*}
    \hat x_{EGL} = \argmax{x\in\mathcal{D}_\mathcal{X}^{(pool)}} \left\{\, \sum\limits_{k = 1}^{K}p_{t,x}(k)
    \norm{ \nabla l_{t}\left(\mathcal{D}_t^{(train)} \cup ( x, k) \right) } \, \right\}.
\end{equation*}
    For the sake of time-complexity we approximate $\hat x_{EGL}$ by 
\begin{equation*}
    \tilde x_{EGL} = \argmax{x\in\mathcal{D}_\mathcal{X}^{(pool)}} \left\{\, \sum\limits_{k = 1}^{K}p_{t,x}(k)
    \norm{ \nabla l_{t}\left(x, k \right) } \, \right\},
\end{equation*}
as $||\nabla l_{t}(\mathcal{D}_t^{(train)})|| \approx 0$ after training $h_t$ ($l_t$ reaches local minimum).

\paragraph*{Impact on the model performance}
The idea is to query instances that can reduce the generalization error. Empirically, we can reduce this error by estimating it directly (e.g. \textit{Expected Error Reduction}) or indirectly by reducing the variance present in the risk of a learning model.
    
The Expected Error Reduction (EER), proposed by \cite{Roy2001TowardOA}, is a strategy of choosing the instance that minimizes the \textit{expected} generalization error since the true label of the instance is currently unknown. We denote $h_{t}^{(x,k)}$ the updated predictor re-trained on $\mathcal{D}_t^{(train)}\cup (x,k)$ at time $t$; $p_{t, u}^{(x,k)}(v)=\mathbb{P}(h_{t}^{(x,k)}(u) = v|u)$ the probability that the class of $u$ is $v$ under $h_{t}^{(x,k)}$ and $y^{(K)} = \argmax{y\in \cal{Y}} \ p_{t, u}^{(x,k)}(y)$.
\cite{Roy2001TowardOA} proposed the following approach for minimizing the expected error (based on 0-1 loss):
\begin{equation*}
    \hat x_{EER} = \argmin{x\in\mathcal{D}_\mathcal{X}^{(pool)}} \left\{ \sum\limits_{u\in\mathcal{D}_\mathcal{X}^{(pool)}}\mathbb{E}_Y \left[1 - p_{t, u}^{(x,Y)}(y^{(K)}) \right] \right\}.
\end{equation*}
The main drawback of this method is its time-complexity. Indeed this sampling method involves $\left|\mathcal{D}_\mathcal{X}^{(pool)}\right|\times K$ re-training of $h_t$. Recently, \citep{mussmann2022active} proposes to reduce this time-complexity by adapting this formulation in a Bayesian perspective for neural networks.

All the sampling methods presented so far prioritize selecting informative instances for the learning model, based on local (cf. Section~\ref{subsec:ALuncer} and ~\ref{subsec:ALdisag}) or global (cf. Section ~\ref{subsec:ALimpac})  prediction  quality. However, some instances may not be representative of the distribution of $\mathcal{X}$, negatively affecting model performance. As an example, anomalies such as \textit{outliers} can be considered informative as they 1) may be in the area of model uncertainty or 2) result in a significant impact on the model parameters after it is labeled. To address this, we suggest considering the representativeness of queried data.

\subsubsection{Sampling based on representativeness}
\label{subsec:ALrepre}

Since an informativeness component alone may not be sufficient, the approaches presented here propose to add a representativeness component: an instance must thus verify a good trade-off between informativeness and representativeness. An example of this methodology is \textit{Information Density} (ID) \cite{SettlesCraven08} where a representativeness component is given by:
\begin{equation*}
\label{eq:ID}
    I_R(x) = \frac{1}{|\mathcal{D}_\mathcal{X}^{(pool)}|}\sum\limits_{u\in\mathcal{D}_\mathcal{X}^{(pool)}} \rm{sim}(x, u) \quad \text{for } \ x\in\mathcal{X}
\end{equation*}
with $\rm{sim}$ a similarity measure between two instances (e.g. the cosine similarity). $I_R(x)$ measures the average similarity between the instance $x$ and the instances of the set $\mathcal{D}_\mathcal{X}^{(pool)}$.

\paragraph{Information Density} We denote $I_A$ an informative importance score such as uncertainty sampling. ID is a density weighted method proposed by \cite{SettlesCraven08} relying on both $I_R$ and $I_A$:
\begin{equation*}
    \hat x_{ID} = \left(\argmax{ x\in\mathcal{D}_\mathcal{X}^{(pool)} } I_A(x, h)\right)\cdot I_R(x)^\beta
\end{equation*}
with $\beta$ a parameter tuning the importance given to the representativeness measure.

There are many other active learning methods that take into account the representativeness of the instances, see \cite{6747346, Ozan18, wang2019bounding}. \cite{wang2019bounding} optimizes for both informativeness and representativeness by solving a single optimization problem and querying multiple instances at each AL update. \cite{Ozan18} suggests finding a representative subset of labeled data (referred to as a \textit{core-set}) that makes a model competitive on the entire dataset by identifying centroids with minimal distance to the rest of the unlabeled data. Nonetheless, this technique necessitates computing a large distance matrix, resulting in significant computational complexity.

\paragraph*{Important remark} Most sampling strategies presented in this Section~\ref{subsec:sampling} can be applied to any off-the-shelf probabilistic model (i.e. model-agnostic). In the following Section~\ref{subsec:ALdeepl}, we present some recent \textit{model-specific} active learning methods based on deep learning and reinforcement learning models respectively. These latter methods are not studied numerically since our fairness component presented later in Section~\ref{sec:FairAL} works only with model-agnostic sampling approaches.

\subsubsection{Other sampling strategies}
\label{subsec:ALdeepl}

We present model-specific AL methods, including deep and reinforcement learning approaches.

\paragraph*{Sampling based on neural nets} Our goal is to estimate uncertainty scores for each unlabeled instance. However deep learning methods pose additional issues in traditional AL setups:
\begin{enumerate}
    \item Active learning methods rely on training models on a small amount of labeled data whereas recent deep learning algorithms are increasingly greedy in terms of labeled data due to the explosion of the number of parameters. Therefore, a too complex neural network architecture can overfit to a "simple" data. Moreover, the \textit{cold-start} problem stated in Section~\ref{subsec:ALuncer} may intensify this issue and consequently may render the previous sampling methods unusable.
    
    \item Most active queries are based on the uncertainty given by the model. However, standard deep learning algorithms do not capture well uncertainties. Indeed, in multi-class tasks, the probabilities obtained with a \textit{softmax} output layer 
    are often misinterpreted as the certainty in the model.  \cite{Gal16Dropout} 
    shows that the \textit{softmax} function results in extrapolations with unjustifiably high certainty for points far from the training data.
\end{enumerate}
Nevertheless, Bayesian methods, such as \textit{Bayesian Deep Learning}, can estimate uncertainties in neural networks~\cite{gal2015dropout}. For instance, \textit{Monte Carlo Dropout} (MC-Dropout)~\cite{Gal16Dropout} is a widely used Bayesian approach that estimates uncertainty as the variance of multiple predictions with dropout filters activated during the inference phase. Another non-Bayesian method proposed by \cite{Lakshminarayanan17} involves training an ensemble of neural networks with the same architecture and different random weights to estimate uncertainty by computing the average of the softmax vectors.

\paragraph*{Sampling based on reinforcement learning}
\label{subsec:ALrl}

In \textit{reinforcement learning} (RL), an agent learns to make decisions by trial-and-error interactions with an environment (see~\citep{wiering2012reinforcement, sutton2018reinforcement}). The association of RL and AL involves using RL to guide the sampling of instances for AL. The idea of this association is to replace hand-crafted rules or heuristics to sample instances for labeling. In the offline scenario, \citep{liu2018learning} considers a Markov decision process in which AL corresponds to the decision to select the most informative instances of $\mathcal{D}_\mathcal{X}^{(pool)}$: the AL strategy is first learned on simulations (e.g., data in which the label of a part of the data is hidden and then revealed to have an automatic oracle) and it is then applied to real AL scenarios. 
Other research, such as \citep{bachman2017learning} or \citep{pang2018meta} use (deep-) RL not only to learn sampling strategies, but also to calibrate the agent so as to generalize across different datasets.

\subsection{Practical considerations}

Active learning methodologies have made theoretical progress but their feasibility in practice remains a challenge. Experimental active learning is difficult for researchers due to lack of access to a labeling oracle. To evaluate these methods, an often-used trick is to transform a labeled database into a suitable one for active learning by masking the label of data not queried. However, this approach raises important issues in practice.

\begin{itemize}
    \item \textit{How to start labeling?} 
    The challenge of \textit{cold-start} is having enough representative data for the learning model before active sampling. Poorly calibrated models may select less informative instances than passive sampling. For initialization, representative instances can be sampled first, followed by uncertainty-based sampling to improve learning. \cite{Sivaraman2011ActiveLF} and \cite{Reitmaier13} propose this approach.
  
    \item \textit{How to evaluate active learning methodologies?} In practice, comparing active learning approaches is challenging with limited budget since they can only be executed once, making exhaustive implementation wasteful. Additionally, budget may be allocated for a hold-out set. To be close to the underlying distribution, it is recommended to first randomly label the hold-out set before considering the training-set.
    
    \item \textit{When to stop the active learning process?} 
    Apart from business considerations such as budget or labeling time, there are various stopping criteria for active learning. For instance, \cite{zhu2007active} proposes the \textit{Max-confidence} and \textit{Min-error} stopping criteria. The former analyzes the entropy of the pool-set, stopping when the entropy of each unlabeled example falls below a certain threshold (e.g. 0.001). The latter examines learning model performance, stopping iterations once a specified level of performance is reached.

    \item \textit{Which learning models to use?} 
    In active learning, the majority of queried samples are tailored to a specific learning model. However, due to limited representative and abundant labeled data initially, selecting an appropriate learning model is challenging. A well-fitted model at the start may not be so by the end of the process.
   
    \item \textit{What happens when labels are unreliable (incorrect or biased labeling)?} When labels are unreliable due to incorrect or biased labeling, the common practice is to relabel uncertain data. If the objectivity of the target variable $Y$ is untrustworthy, especially in cases of social or historical biases, the methods introduced later in Section~\ref{sec:FairAL} can be employed to enforce Demographic Parity fairness.

\end{itemize}

\section{Numerical illustration on synthetic data}
\label{subsec:illustration}

In this section we discuss several numerical aspects of the studied procedure on synthetic datasets. Passive learning is used as a benchmark. In a nutshell we illustrate the efficiency and behaviour of different active learning processes to build an accurate (in Section \ref{subsubsec:illustration_perf}) and fair (in Section \ref{subsubsec:illustration_unf}) logistic regression model. As such, we study the effectiveness of some traditional (model-agnostic) sampling strategies: Entropy, QBag, EGL and ID.

\subsection{Model performance}
\label{subsubsec:illustration_perf}





\begin{figure}[H]
\begin{center}
\includegraphics[scale=0.38]{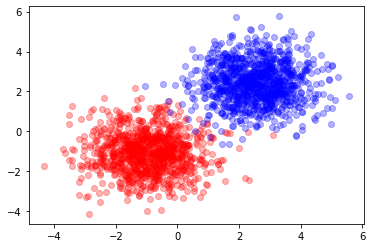}
\includegraphics[scale=0.5]{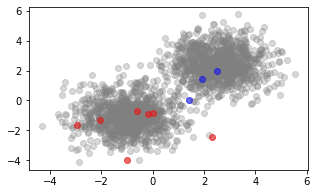}
 \caption{Synthetic dataset: \texttt{Two-Gaussians}.}
Left: fully labeled dataset.  \quad Right: pool-set (gray points) and train-set (red or blue points).
\label{fig:toy_example}
\end{center}
\end{figure}

\paragraph*{Synthetic dataset} All studies in this section are realized on a Gaussian dataset called \texttt{Two-Gaussians}. We have generated 2000 Gaussian instances of dimension 2: 1000 \textit{red} examples and 1000 \textit{blue} examples with same variance but different means. These examples will then constitute the set of labeled data (red or blue points $
\{\tikz\draw[red,fill=red] (0,0) circle (.5ex);, \tikz\draw[blue,fill=blue] (0,0) circle (.5ex);\}$) and unlabeled data (gray points $
\{\tikz\draw[gray,fill=gray] (0,0) circle (.5ex);\}$). At initialization (iteration 0 of AL) 10 instances are randomly queried for constructing the first training-set (see Figure~\ref{fig:toy_example}). Note that the metric accuracy (denoted \textit{acc}, which corresponds to the complementary of the misclassification risk) is used for comparing the models performances since the synthetic dataset is balanced.

\begin{figure}[H]
\begin{center}
\includegraphics[scale=0.41]{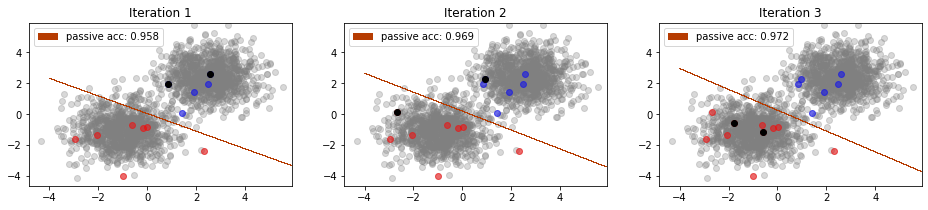}
\caption{Passive learning - first 3 iterations of random sampling of two instances (in black).}
\label{fig:PL_gauss}
\end{center}

\end{figure}

\begin{figure}[H]
\begin{center}
\includegraphics[scale=0.41]{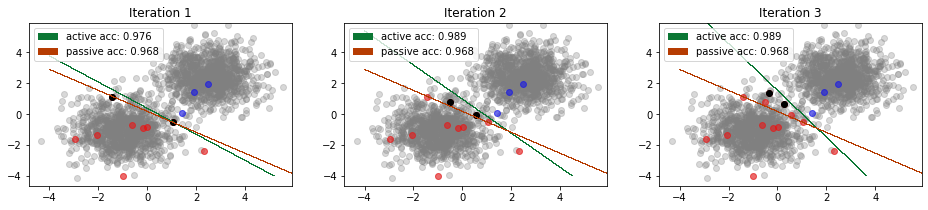}
\caption{Sampling by Shannon Entropy.}
\label{fig:ALent_gauss}
\end{center}
\end{figure}

\begin{figure}[H]
\begin{center}
\includegraphics[scale=0.41]{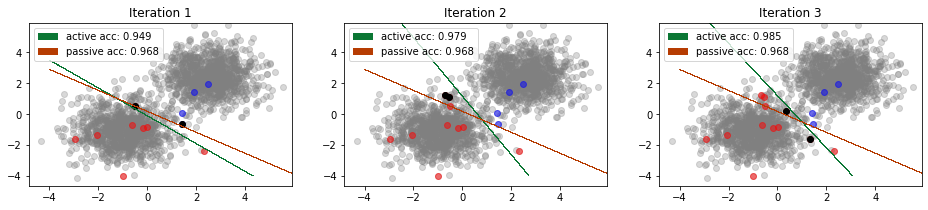}
\caption{Sampling by Query by Bagging.}
\label{fig:ALqbag_gauss}
\end{center}
Disagreement measure: \textbf{vote entropy}; Model committee: 8 logistic models (\textbf{query-by-bagging}).
\end{figure}

\begin{figure}[H]
\begin{center}
\includegraphics[scale=0.41]{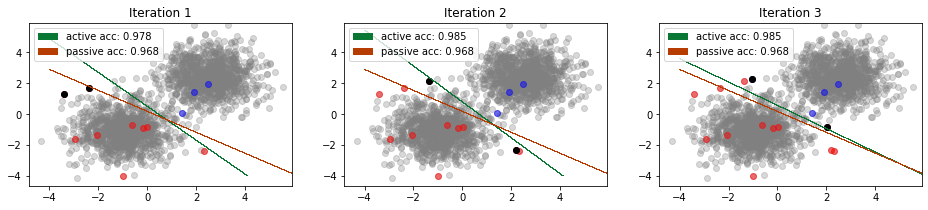}
\caption{Sampling by Expected Gradient Length.} 
\label{fig:ALegl_gauss}
\end{center}
\end{figure}

\begin{figure}[H]
\begin{center}
\includegraphics[scale=0.41]{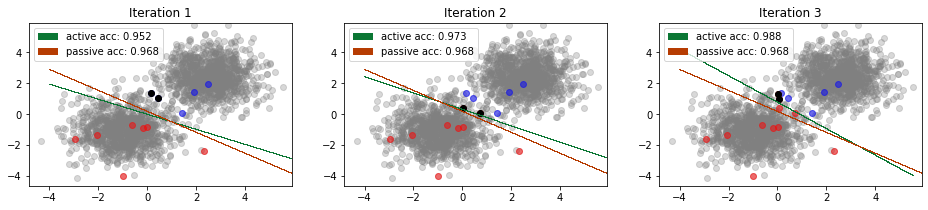}
\caption{Sampling by Information Density (informativeness score: Shannon entropy).} 
\label{fig:ALdensity_gauss}
\end{center}
\end{figure}

\paragraph*{Results} Figure \ref{fig:PL_gauss} displays the behaviour of passive learning on the synthetic dataset. This figure is used as a benchmark. After 3 iterations, passive learning updates slowly (the estimator keeping roughly the same accuracy) whereas Figures \ref{fig:ALent_gauss}, \ref{fig:ALqbag_gauss}, \ref{fig:ALegl_gauss} and \ref{fig:ALdensity_gauss} illustrate how active learning outperforms the classical sampling way. The \textit{green line} illustrates the decision threshold of the logistic model after training on the labeled data and the \textit{red line} corresponds to the passive model in Figure~\ref{fig:PL_gauss}.
Uncertainty and disagreement based sampling (Figures \ref{fig:ALent_gauss} and \ref{fig:ALqbag_gauss}) focus on the model uncertainty to classify. The loss-based sampling (Figure \ref{fig:ALegl_gauss}) focus more on the \textit{non-dense} tails of the two normal distributions (near model uncertainty regions but far from dense regions). At last, the density-based sampling (Figure \ref{fig:ALdensity_gauss}) queries both uncertain and representative (near dense regions) instances.

%

Up to now, we have evaluated and compared active learning methods based on their ability to improve model performance (e.g. misclassification risk or, equivalently, accuracy) by labeling informative and/or representative instances. Let us now consider in the next section the \textit{fairness} of these methodologies.

\subsection{Model unfairness}
\label{subsubsec:illustration_unf}

\begin{figure}[H]
\begin{center}
\includegraphics[scale=0.38]{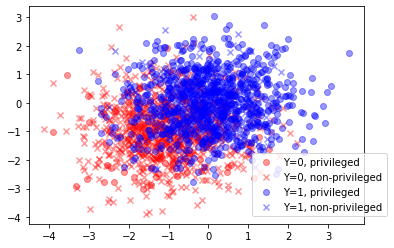}
\includegraphics[scale=0.5]{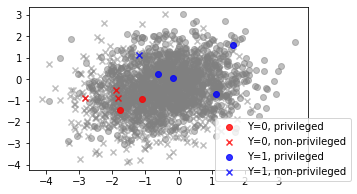}
 \caption{Synthetic dataset: \texttt{Two-Gaussians-Unfair}.}
Left: fully labeled dataset.  \quad Right: pool-set (gray points) and train-set (red or blue points).
\label{fig:TWO_GAUSSIAN_UNFAIR}
\end{center}
\end{figure}

Following the previous synthetic dataset \texttt{Two-Gaussians}\footnote{Our experimentation can be found at \url{https://github.com/curiousML/active-sampling/blob/master/notebooks/gaussians_analysis.ipynb}.} (with different means) where the class and its availability are presented by colored points \{\tikz\draw[gray,fill=gray] (0,0) circle (.6ex);, \tikz\draw[red,fill=red] (0,0) circle (.6ex);, \tikz\draw[blue,fill=blue] (0,0) circle (.6ex);\} or $Y \in \{\text{NA}, 0, 1\}$, we introduce (with the same color rule) the privileged group by \{\tikz\draw[gray,fill=gray] (0,0) circle (.5ex);, \tikz\draw[red,fill=red] (0,0) circle (.5ex);, \tikz\draw[blue,fill=blue] (0,0) circle (.5ex);\} and the non-privileged group by $\{\textcolor{gray}{\boldsymbol{\times}}, \textcolor{red}{\boldsymbol{\times}}, \textcolor{blue}{\boldsymbol{\times}}\}$. The unfairness in the dataset is denoted by a new binary sensitive feature $S\in \{ -1, 1 \} = \{\tikz\draw[black] (0,0) circle (.6ex);, \times\}$. $S$ is assigned based on the label, $Y$, with $\left(S | Y = 0\right) \sim 2\cdot \mathcal{B}(p) - 1$ and $\left(S | Y = 1\right) \sim 2\cdot \mathcal{B}(1- p) - 1$. $\mathcal{B}$ is the Bernoulli distribution and the parameter $p$ measures the level of unfairness in the dataset, with $p=0.5$ indicating fairness and larger deviations from $0.5$ indicating greater unfairness. We set $p=0.9$ as a default value and we call this new synthetic dataset \texttt{Two-Gaussians-Unfair}. With this sensitive feature, this dataset contains instances of dimension 3. See Figure~\ref{fig:TWO_GAUSSIAN_UNFAIR} for an illustration. We denote informally,
$$\text{rate}_{S=s}(y) := \frac{\#\{\text{correct prediction  $\hat{Y}=y$ given $S=s$}\}}{\#\{\text{correct prediction  $\hat{Y}=y$}\}},$$
the proportion of the correct predictions of $\hat{Y}=y$ given the group $S=s$, out of all correct predictions for the same $y$ across all values of $S$. The variable $\hat{Y}$ corresponds to the outcomes of the logistic regression model.
To quantify unfairness in Demographic Parity, Equation \eqref{eq:unfairnessDP} in Section \ref{subsec:fairdef} can be used. In this specific case, unfairness can be measured empirically by
\begin{equation*}
\text{Unf}:= \max\limits_{y\in\{0,1\}} \left| \ \text{rate}_{S=-1}(y) - \text{rate}_{S=1}(y) \ \right|.
\end{equation*}

\begin{figure}[H]
\begin{center}
\includegraphics[scale=0.4]{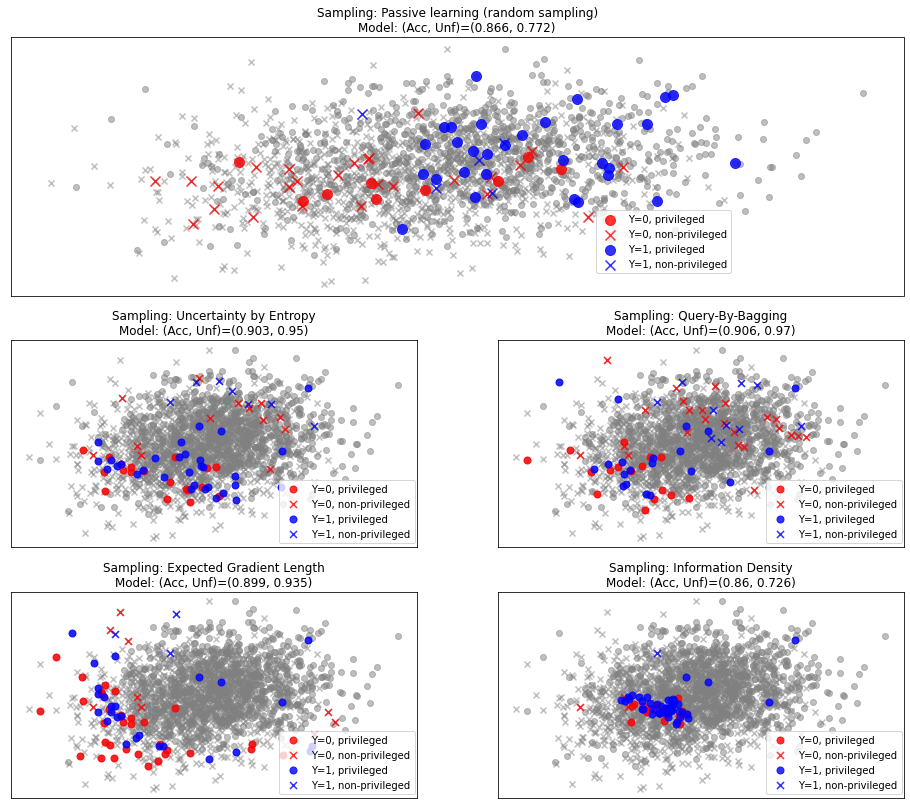}
\caption{Passive and active sampling on synthetic data with fairness study.} 
\label{fig:UNFAIRNESS_TWO_GAUSSIAN}
\end{center}
\end{figure}

 Figure~\ref{fig:UNFAIRNESS_TWO_GAUSSIAN} illustrates the performance comparison of passive and active learning methods on the \texttt{Two-Gaussians-Unfair} dataset. The experiment was conducted for 20 iterations, where three instances were sampled per iteration. The figure indicates that active learning tends to select informative samples from underrepresented non-privileged groups in privileged groups and vice versa, which balances the training-set. However, this approach may lead to accentuated unfairness in the hold-out set. This is because the model maximizes its predictive performance by relying on characteristics that are correlated with group membership, which may perpetuate unfairness and reproduce biases in the hold-out set.



\begin{table*}[h!]
\begin{tiny}
    \resizebox{\linewidth}{!}{
    \begin{tabular}{|l||*{4}{c|}} \hline
    \multirow{2}{*}{\begin{tabular}{p{1.5cm}}\backslashbox[26mm]{SAMPLING}{MODEL} \end{tabular}
    } & \multicolumn{2}{c|}{Base model $h$} \\ \cline{2-3}
     & Accuracy & Unfairness \\ \hline\hline
    Random sampling (PL) & 
    $0.866$ &
    $0.772$ \\ 
    Entropy sampling (AL) &
    $0.903$  & 
    $0.95$ \\
    Qbag sampling (AL) & 
    $0.906$ & 
    $0.97$ \\
    EGL sampling (AL) &
    $0.899$ & 
    $0.935$ \\
    ID sampling (AL) &
    $0.86$ & 
    $0.726$ \\
    \hline
    \end{tabular}}
    \caption{Accuracy \& Unfairness of the sampling methods for synthetic data. When comparing AL methods to the PL method with the same sample size (though not identical sets), most AL approaches improve accuracy as expected but at the expense of increased unfairness. 
    }
    \label{table:res_gauss_base}
    \end{tiny}
\end{table*}

Table~\ref{table:res_gauss_base} presents numerical results indicating that many traditional active learning approaches, except ID sampling (unfairness of 0.726), can lead to increased unfairness (with unfairness score of over 0.9), compared to passive learning's score of 0.772.  Indeed, traditional active learning methods can only identify informative instances and are inadequate in handling model unfairness, especially when it is caused by label bias. To achieve fairer instance sampling without compromising predictive performance, fairness reduction must be explicitly incorporated into sampling strategies. In the next section we propose a novel approach called \textit{model-agnostic fair active learning} (or \textit{fair active learning} in short) that both  enhances the fairness and  allows fairer instance sampling in any model-agnostic sampling strategy.

\section{Fair active learning}
\label{sec:FairAL}

To make our sampling fair w.r.t. a sensitive characteristic (such as gender), we need to find a good compromise between the misclassification risk $R$ of  Equation~\eqref{eq:defR} and the unfairness $\mathcal{U}$ of the Equation~\eqref{eq:unfairnessDP}.  This is achieved by the constrained optimization proposed in \cite{denis2024fairness} which achieves state-of-the-art unfairness (under Demographic Parity) results in both binary and multi-class tasks.

\subsection{Fair multi-class classification} 
\cite{denis2024fairness} provides a  post-processing method to add Demographic Parity constraints in multi-class classification
(see Section~\ref{subsec:fairdef}) with theoretical guarantees both in terms of fairness and risk. The associated plug-in estimator is also provided. In our study, we denote this estimator $h_t^{(fair)}$.
This post-processing technique offers two significant advantages. Firstly, it can be utilized with any off-the-shelf classifier. Secondly, it requires a pool of unlabeled data to compute certain quantities, and the larger the pool, the better the unfairness guarantee. This makes it appropriate for the offline scenario.

Let us denote $(p_{t, (x, s)}(k))_{k\in[K]}$ the conditional probabilities (e.g. logistic regression, random forest, \emph{etc}.) at time $t$ (for the instance $x$ with the associated sensitive feature $s$) trained on the training-set $\mathcal{D}_t^{(train)}$. We also denote 
$(\zeta_{k})_{k\in [K]}$ and $(\zeta_{k,i}^s)$ as $K$ and $K\times |\mathcal{D}_\mathcal{X}^{(pool)}|$ i.i.d samples from a uniform distribution on $[0,u]$ (e.g. $u=10^{-5}$). Following \cite{denis2024fairness}, and readjusting the result for exact fairness, the new scores are given by
\begin{equation}
\label{eq:eqPlugIn}
p^{(fair)}_{t, (x, s)}(k) = \hat{\pi}_s (p_{t, (x, s)}(k)+\zeta_k) - s \hat{\lambda}_k\, ,\;\; \mbox{for all }(x,s)\in \mathcal{X}_{-S}\times {\cal{S}}
\end{equation}
with $\hat{\lambda} \in \mathbb{R}^K$ given as
\begin{equation*}
\hat{\lambda} = \argmin{\lambda} \sum_{s \in \mathcal{S}}
\frac{1}{N_s} \sum_{i=1}^{N_s} \left[\max_{k\in [K]} \left(\hat{\pi}_s (p^s_{t, X_i^{s}}(k)+\zeta^s_{k,i}) -s\lambda_k\right)\right]
\end{equation*}
and where (in a semi-supervised way) we use the pool-set $\mathcal{D}_\mathcal{X}^{(pool)}$ to calibrate:
\begin{itemize}
    \item $(\hat{\pi}_s)_{s\in\mathcal{S}}$ the empirical frequencies for estimating the distribution of the sensitive feature $\mathcal{S}$;
    \item $N_s$ the number of observations corresponding to $\{S=s\}$. Therefore, $N_{-1}+N_{1} = \left| \mathcal{D}_\mathcal{X}^{(pool)} \right|$;
    \item $X_1^s, \ldots , X_{N_s}^s \subset \mathcal{D}_\mathcal{X}^{(pool)}$ composed of \emph{i.i.d.} data from the distribution of $X^s:= \{X|S=s\}$.
\end{itemize}
The associated optimal (fair) prediction is:

\begin{equation}
\label{eq:PlugInArgmax}
{h}_t^{(fair)}(x, s) = \argmax{k\in [K]} p^{(fair)}_{t, (x, s)}(k).
\end{equation}

The training-set $\mathcal{D}_t^{(train)}$ and the pool-set $\mathcal{D}_\mathcal{X}^{(pool)}$ are both used to generate the new scores $p^{(fair)}_{t, (x, s)}(k)$.




\subsection{Fair active sampling} 
\begin{figure}
\begin{center}
\includegraphics[scale=0.36]{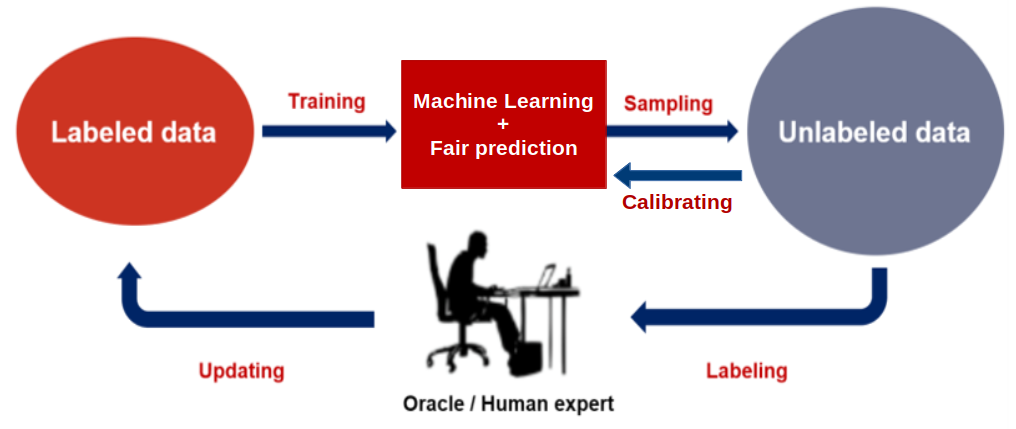}
\caption{Fair active learning in an offline scenario - to calibrate the fair prediction in a multi-class setting, we use both labeled and unlabeled data (semi-supervised approach).}
\label{fig:FAL}
\end{center}
\end{figure}

In model-agnostic active learning, choosing the best sample that balance risk and fairness can be achieved by using the optimal fair prediction ${h}_t^{(fair)}$ instead of the (unfair) prediction $h_t$ in the maximisation of the importance score function. This model-agnostic \textit{fair active learning} (FAL) approach 
is presented in Algorithm~\ref{alg:fairactivelearning} and illustrated in Figure \ref{fig:FAL}. Since we render the labeled data $\mathcal{D}_t^{(train)}$ fairer w.r.t. a sensitive feature, the estimator $h_t$ trained on this data should also be fair. In particular, this approach can be applied to all sampling strategies that use the output of the probabilistic learning model (also called \textit{query-agnostic}).  A comparison between different methods is provided in Table \ref{Table:comparisonmodel}.

\begin{algorithm}
   \caption{Model-agnostic fair active learning}
   \label{alg:fairactivelearning}
\begin{algorithmic}
   \STATE {\bfseries Input:} $h$ a base estimator, $\mathcal{D}^{(train)}$ the initial training-set and $\mathcal{D}_\mathcal{X}^{(pool)}$ the initial pool-set.\medskip
   
   \STATE {\bf \quad Step 0.} Fit $h$ on the training-set $\mathcal{D}^{(train)}$.\medskip

    \STATE {\bf \quad Step 1.} Compute $p^{(fair)}_{(x, s)}(k)$  using Equation~\eqref{eq:eqPlugIn}
     \\  \hfill (\texttt{Accelerated gradient descent} can be used for the optimization, see~\citep{Nesterov83,Nesterov13}).
    \medskip
    
    \STATE {\bf \quad Step 2.} Compute ${h}^{(fair)}(x, s)$ using Equation~\eqref{eq:PlugInArgmax}\medskip
    
   \STATE {\bf \quad Step 3.} The fair active learner queries the instance $x^{*} \in \mathcal{D}_\mathcal{X}^{(pool)}$ that maximizes the importance score function $I(x, {h}^{(fair)})$ i.e.
   
    \begin{equation*}
        x^{*} = \argmax{x \in \mathcal{D}_\mathcal{X}^{(pool)}} \left\{ I(x, {h}^{(fair)}) \right\}
    \end{equation*}
   
   \STATE {\bf \quad Step 4.} Update the training-set and the pool-set: if we denote $y^{*}$ its label then 
   \begin{eqnarray*}
       \mathcal{D}^{(train)} & = & \mathcal{D}^{(train)} \cup \{(x^{*}, y^{*})\}\\
       \mathcal{D}_\mathcal{X}^{(pool)} & = & \mathcal{D}_\mathcal{X}^{(pool)} - \{x^{*}\}
   \end{eqnarray*}
   
   \STATE {\bf \quad Step 5.} As long as we do not reach a stopping criterion (e.g. exhaustion of the labeling budget or convergence of the performance), we repeat this process (\textbf{return to step 0}).\medskip
   \STATE {\bfseries Output:} the final estimators $h$ and ${h}^{(fair)}$; the \textit{fairer} dataset $\mathcal{D}^{(train)}$.
\end{algorithmic}
\end{algorithm}

\begin{table}
\label{sample-table}
\vskip 0.15in
\begin{center}
\begin{scriptsize}
\begin{sc}
\begin{tabular}{lccccr}
\toprule
Method & Multi-class & Fair (DP) &  Model-agnostic & Query-agnostic\\
\midrule
\midrule
PANDA \cite{sharaf2022promoting} & NA & $\surd$ & $\times$ & $\times$\\\midrule
PL & $\surd$ & $\times$ & $\surd$ & $\surd$\\\midrule
AL (Alg.~\ref{alg:activelearning}) & $\surd$ & $\times$ & $\surd$ & $\surd$\\
\midrule
\textbf{FAL (Alg.~\ref{alg:fairactivelearning})}     & $\surd$ & $\surd$ &    $\surd$ & $\surd$     \\
\bottomrule
\end{tabular}
\end{sc}
\end{scriptsize}
\end{center}
\caption{Comparison between methods. PANDA is a fair active learning method for binary tasks.
\textbf{Multi-class}: the implementation can be applied to multi-class tasks; \textbf{Fair}: the method enforces fairness under DP; \textbf{Model-agnostic}: the implementation can be applied to any probabilistic model; \textbf{Query-agnostic}: the method can be applied to any query using probabilistic outputs.}
\label{Table:comparisonmodel}
\end{table}

\subsection{Results on unfair synthetic data}

\begin{figure}[H]
\begin{center}
\includegraphics[scale=0.4]{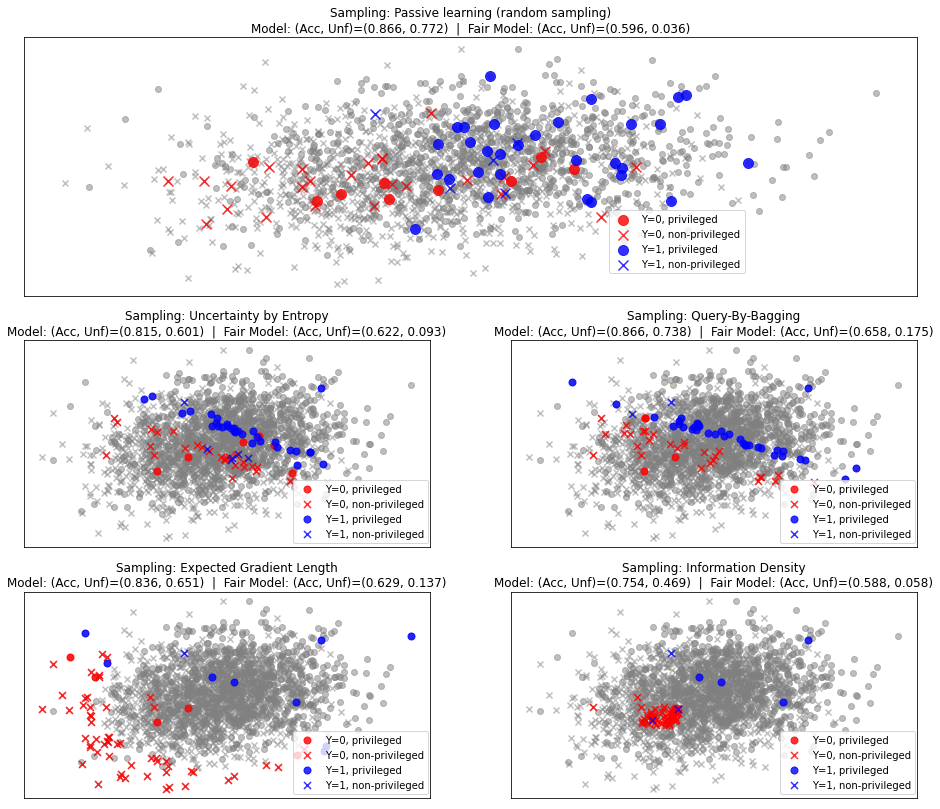}
\caption{Passive and \textbf{fair} active sampling on synthetic data with fairness study.} 
\label{fig:UNFAIRNESS_TWO_GAUSSIAN_OPTIMAL}
\end{center}
\end{figure}

We apply our fair active learning methods on the synthetic dataset \texttt{Two-Gaussians-Unfair} presented in Section~\ref{subsubsec:illustration_unf}. Figure~\ref{fig:UNFAIRNESS_TWO_GAUSSIAN_OPTIMAL} illustrates the behaviour of active and passive learning on this data. The prediction made by our fair model is based on all the features except the sensitive feature (as the model is enforced to be independent), which allows our model to achieve Demographic Parity. This means that the importance score in FAL is decorrelated to the sensitive feature. This leads to a fairer selection compared to passive learning. Table~\ref{table:res_gauss_optimal} indicates the efficacy of our fair active learning approach via two models: the (unmodified) base model $h$ and the (modified) fairer model $h^{(fair)}$. The outputs of the base model trained on fair instances demonstrates its effectiveness in selecting more fair instances while maintaining competitive predictive performance. Meanwhile, the outputs of the fairer model highlights the capability of our proposed approach to achieve exact-fairness, independent of the sample selected.


\begin{table*}[h!]
\begin{tiny}
    \resizebox{\linewidth}{!}{
    \begin{tabular}{|l||*{6}{c|}} \hline
    \multirow{2}{*}{\begin{tabular}{p{1.5cm}}\backslashbox[26mm]{SAMPLING}{MODEL} \end{tabular}
    } & \multicolumn{2}{c|}{Base model $h$} & \multicolumn{2}{c|}{Fair model $h^{(\text{fair})}$} \\ \cline{2-5}
     & Accuracy & Unfairness & Accuracy & Unfairness \\ \hline\hline
    Random sampling (PL) & 
    $0.866$ &
    $0.772$ &
    \cellcolor{gray!25} & 
    \cellcolor{gray!25} \\ 
    Entropy sampling (AL) &
    $0.903$  & 
    $0.95$ & 
    \cellcolor{gray!25} & 
    \cellcolor{gray!25} \\
    Qbag sampling (AL) & 
    $0.906$ & 
    $0.97$ & 
    \cellcolor{gray!25} & 
    \cellcolor{gray!25} \\
    EGL sampling (AL) &
    $0.899$ & 
    $0.935$ & 
    \cellcolor{gray!25} & 
    \cellcolor{gray!25} \\
    ID sampling (AL) &
    $0.86$ & 
    $0.726$ & 
    \cellcolor{gray!25} & 
     \cellcolor{gray!25}\\ \hline
     FairEntropy sampling (FAL) &
    $0.815$ & 
    $0.601$ & 
    $0.622$ & 
    $0.093$ \\
    FairQbag sampling (FAL) & 
    $0.866$ & 
    $0.738$ & 
    $0.658$ & 
    $0.175$ \\
    FairEGL sampling (FAL) &
    $0.836$ & 
    $0.651$ & 
    $0.629$ & 
    $0.137$ \\
    FairID sampling (FAL) &
    $0.754$ & 
    $0.469$ & 
    $0.588$ & 
    $0.058$ \\
    \hline
    \end{tabular}}
    \caption{Accuracy \& Unfairness of the sampling methods for synthetic data. When comparing AL methods to the PL method with the same sample size (though not identical sets), most AL approaches improve accuracy as expected but at the expense of increased unfairness. In remedy, FAL consistently maintains lower unfairness levels, with a slight decrease in predictive performance using model $h$. For a more constrained approach, we alternatively propose the model $h^{(fair)}$, which optimizes accuracy under the exact DP fairness constraint, thus validating the near-zero unfairness metric. 
    }
    \label{table:res_gauss_optimal}
    \end{tiny}
\end{table*}

\section{Actuarial case study}\label{sec:alemp}

We now present the empirical effectiveness of model-agnostic AL (Algorithm~\ref{alg:activelearning}) and FAL (Algorithm~\ref{alg:fairactivelearning}) approaches on (real) insurance datasets: an insurance case presented in Section~\ref{subsec:datainsurance} and a sentiment analysis on textual data from an insurance dataset in Appendix~\ref{appsec:addfigAL}. 

\subsection{Metrics}
\label{subsec:metrics}

To evaluate the model performance and the model unfairness we consider two metrics: the \textit{$F_1$-score} and the \textit{empirical unfairness measure}.

\paragraph*{Performance measure} The well-known \textit{accuracy} (e.g. empirical form of the complement of the misclassification risk) is not used due to potential misinterpretation in imbalanced datasets: a very high accuracy might not represent a good model (e.g. a constant classifier always predicting the majority class). In binary task, we use $F_1$-score instead: the harmonic mean of \textit{precision} (i.e. number of true positives divided by the total number of predicted positives) and \textit{recall} (i.e. number of true positives divided by the total number of actual positives),

\begin{equation*}
\label{eq:ModelMetricEmp}
    F_1 = 2\times \frac{\rm{precision}\times \rm{recall}}{\rm{precision}+\rm{recall}}.
\end{equation*}

The $F_1$-score has a good balance between precision and recall and is therefore a good metric for balanced and imbalanced datasets. For a multi-class task we can compute one $F_1$-score per class in a one-vs-rest manner and then average the results. 

\paragraph*{Unfairness measure} The fairness of an estimator $h \in \mathcal{H}$ is measured on a test-set $\mathcal{D}^{(test)}$ via the empirical counterpart of the unfairness measure $\mathcal{U}(h)$ given in Section~\ref{subsec:fairdef}. For simplicity's sake, we note $\hat \nu_{h|s} (k)= \frac{1}{|\mathcal{T}^{s}|}\sum_{(X, Y) \in \mathcal{T}^{s}}\mathds{1}_{\{h(X, s) = k\}}$ the empirical distribution of $h(X, S) | S = s$  on the  conditional test-set $\mathcal{T}^s = \left\{ (X, Y)\in \mathcal{D}^{(test)} | S = s \right\}$. The empirical unfairness measure (or simply unfairness) is formally defined by:

\begin{equation}
\label{eq:ModelUnfairEmp}
    \hat{\mathcal{U}}(h) = \max_{k \in[K]}  \left|
    \hat \nu_{h|-1} (k) - \hat \nu_{h|1} (k) \right|.\enspace
\end{equation}

The metric ($F_1$-score, unfairness) will be used to compare PL, AL and our FAL methods.

\subsection{Methods and settings}

The learning model used is XGBoost (see \cite{Tianqi16}), denoted \texttt{GbmModel}. This algorithm has the advantage of being flexible and is faster to train than Gradient Boosting. Thanks to a regularization term it adapts rather well to small or medium size data. We consider the benchmark passive sampling approach \texttt{RandomSampling} as well as:
\begin{itemize}
    \item the model-agnostic sampling strategies: \texttt{EntropySampling}, the uncertainty sampling with entropy measure; \texttt{QbagSampling}, the disagreement-based sampling with Query-by-bagging; \texttt{EGLSampling}, the model change sampling strategy with EGL and \texttt{DensityWeightedSampling} the density-based sampling;

    \item and the respectively associated fair sampling strategies: \texttt{FairEntropySampling}, \texttt{FairQbagSampling},
\texttt{FairEGLSampling}
and
\texttt{FairDensityWeightedSampling} with respect to a sensitive attribute such as gender.
\end{itemize}

Note that the effectiveness of active learning techniques in improving model performance is clear: in the experiment presented in  Appendix~\ref{appsec:addfigAL} for sentiment analysis with the dataset \textit{Net Promoter Score} of Société Générale Insurance, most strategies reveal needing 5 times less labeled data than passive learning to achieve the same $F_1$-score. In addition, AL is robust to some moderately imbalanced datasets. See Appendix~\ref{appsec:addfigAL} for more details. In the next section, we will examine the insurance dataset \textit{French Motor Personal Line} (\texttt{freMPL}) and evaluate how well AL and FAL\footnote{The source of our method can be found at \url{https://github.com/curiousML/active-sampling}.} methods can sample informative and fair instances.

\subsection{Actuarial dataset}
\label{subsec:datainsurance}

The database \texttt{freMPL}, described in \cite{dutang2020package}, is a collection of data related to car insurance policies in France. The dataset contains around $30,000$ policies for year $2004$ and includes information on the policyholders, such as their age, gender, and geographic location, as well as details about the vehicles covered by the policies, such as the model of the vehicle, the year it was manufactured, and the value of the vehicle. Additionally, the dataset contains information on any claims made on the policies, including the date 
and the amount of the claim.

Our task is to predict if policyholders have a garage for their vehicle, which can be a crucial factor in determining the pricing of their car insurance (see Appendix~\ref{appsec:garage} how the feature `garage' impacts claims). A garage can indicate lower risk for certain types of claims, such as accidents, car theft, and weather-related damage, compared to parking on the street or in an open lot. Thus, the presence of a garage can significantly impact the overall risk profile.

Note that factors such as the garage can be identified 1) manually using tools like \textit{Google Street View} by a human~\cite{kita2019google} or 2) automatically with computer vision~\cite{blier2021rethinking}. Nevertheless, the first approach is too time-consuming and costly to consider, while the second requires a labeled dataset. In practice, in order to better exploit the data and set up a more accurate (and fair) model, one can use AL (FAL). As the feature `garage' is already present in \texttt{freMPL}, we can run AL by masking the labels of certain instances. Gender is used as a sensitive feature for fairness analysis.


\subsection{Results}
\label{subsec:FALresults}

\begin{figure}[H]
\begin{center}
\begin{subfigure}{\textwidth}
 \centering
\includegraphics[scale=0.7]{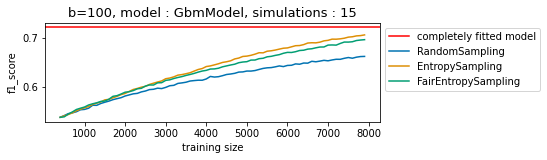}
\caption{}
\label{fig:fairALperf}
\end{subfigure}
\begin{subfigure}{\textwidth}
 \centering
\includegraphics[scale=0.7]{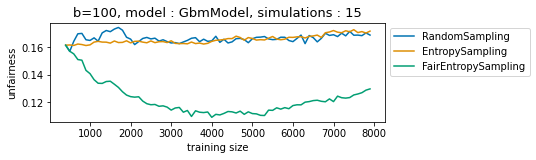}
\caption{}
\label{fig:fairALunf}
\end{subfigure}
\caption{Performance and unfairness of PL, AL and FAL approaches on \texttt{freMPL} dataset. Model evaluation ($F_1$-score and Unfairness) of PL, AL and FAL for the model \textbf{XGBoost} w.r.t the training-set iteratively constructed with the sampling stategies mentionned above. At each iteration, we query $b=100$ instances.}
\label{fig:fairALperfunf}
\end{center}
\end{figure}

\begin{figure}[H]
\begin{center}
\includegraphics[scale=0.51]{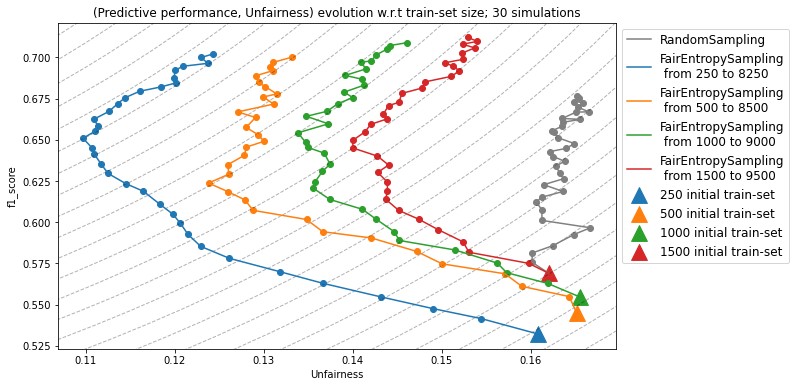}
\caption{($F_1$-score, Unfairness) evolution w.r.t. the training-set. Iterations start at the bottom-right corner of the graph. Top-left corner gives the best trade-off. Due to the finite pool-set, all methodologies involving $h$ ultimately reach the same endpoint regardless of the initial train-set ($\bigtriangleup$) and sampling strategies (---).}
\label{fig:AL_init_sample_analysis}
\end{center}
\end{figure}

\begin{figure}[H]
\begin{center}
\includegraphics[scale=0.285]{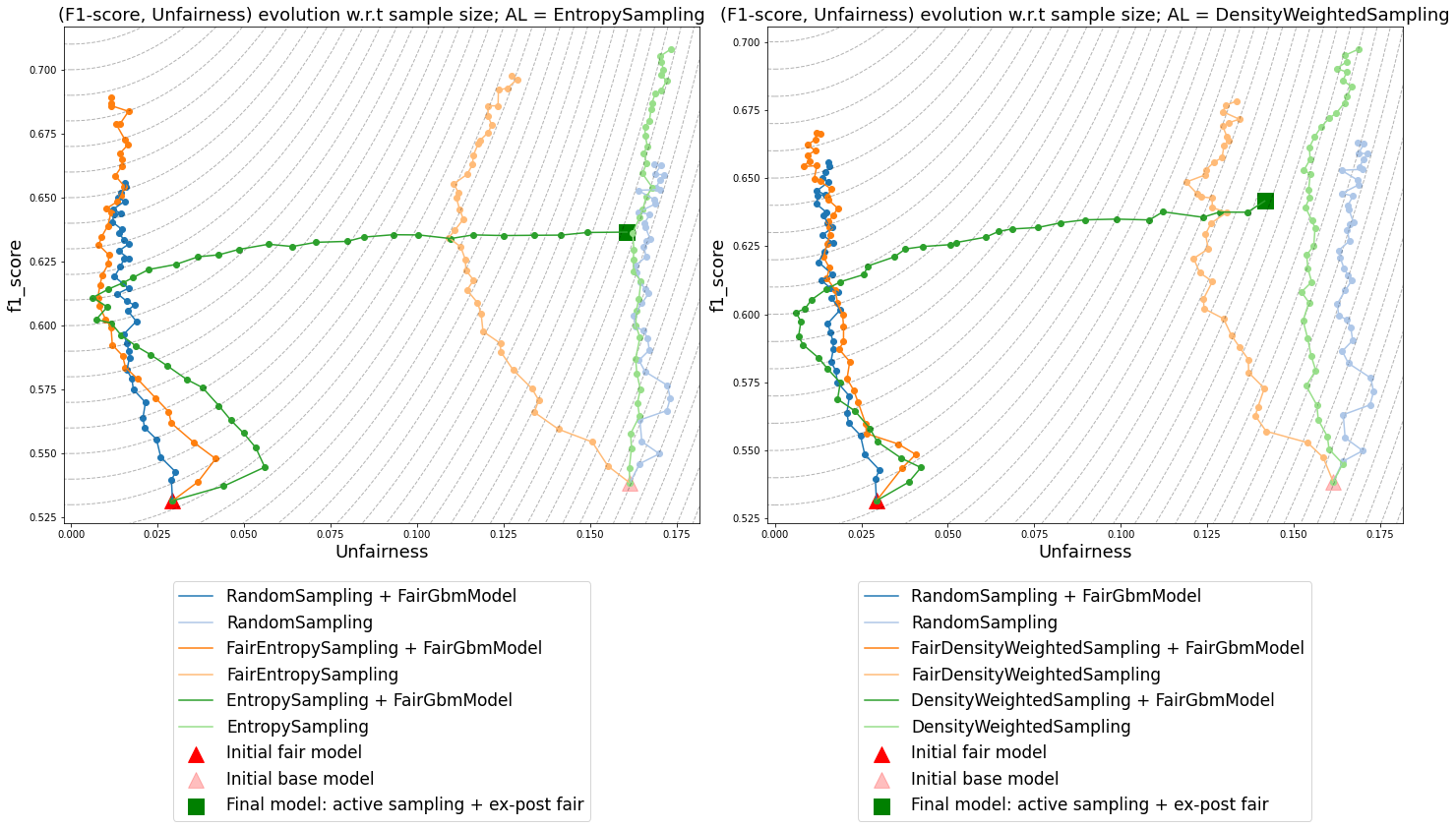}
\caption{($F_1$-score, Unfairness) evolution w.r.t. the training-set. Iterations start at the bottom of the graph. Top-left corner gives the best trade-off. The (non-transparent) green curve shows outcomes of ex-post fairness calibration on (uncalibrated) AL strategies, with each calibration being independent of the employed AL.}
\label{fig:FAL_FML_analysis}
\end{center}
\end{figure}

For simplicity sake, we present in this section the methods \texttt{RandomSampling}, \texttt{EntropySampling} and \texttt{FairEntropySampling}. All the other methods are presented in Appendix~\ref{appsec:add_garage} and show similar tendencies. All the graphs presented are averaged over 15 or 30 simulations, where each simulation corresponds to an initial training-set randomly selected from the pool-set. We further explored the influence of (a) the initial (passive) sample size on AL and (b) the performance of the baseline ex-post unfairness calibration in AL scenarios. In response to this feedback, we carried out additional numerical experiments (refer to Figure~\ref{fig:AL_init_sample_analysis} for (a) and Figure~\ref{fig:FAL_FML_analysis} for (b)) with an increased number of simulations (30 simulations). We also highlighted specific desired properties for FAL, which are explained below. Additionally, note that once the pool-set is exhausted (finite pool-set), all methods reach the same results as the fully fitted model. 

\paragraph*{Fair active learning with base model $h$} We evaluate AL and FAL efficiency compared to PL. More specifically, to assess the informativeness and fairness of the sampling approaches, we evaluate the base model $h$. 
\begin{itemize}
    \item \textbf{model performance} Figure~\ref{fig:fairALperf} demonstrates that both \texttt{EntropySampling} and \texttt{FairEntropySampling} perform better than \texttt{RandomSampling}. Indeed, although FAL has constrained optimization leading to a performance decrease compared to AL, the result remains surprisingly competitive (see Figure~\ref{fig:fairALunfApp} in Appendix~\ref{appsec:add_garage_FALvsPL}).

    \item \textbf{model unfairness} Figure~\ref{fig:fairALunf} reveals that \texttt{EntropySampling} is as unfair as \texttt{RandomSampling} whereas \texttt{FairEntropySampling} strategy, as expected, outperforms both approaches. In particular, Figure~\ref{fig:FALperformance_unfairnessApp1} in Appendix~\ref{appsec:add_garage_FALvsALvsPL} shows that some AL strategies are intrinsically unfair compared to PL, while FAL makes these approaches much more fair.
    
\end{itemize} 
In Figure~\ref{fig:AL_init_sample_analysis}, FAL performance is illustrated for various initial (passively selected) training sizes in $\{250, 500, 1000, 1500\}$, with each FAL strategy labeling an additional $8000$ instances. Notably, the figure suggests that beginning FAL methods with fewer samples can improve both fairness and accuracy (see the blue curve or even the orange curve).

\begin{figure}[H]
\begin{center}
\includegraphics[scale=0.5]{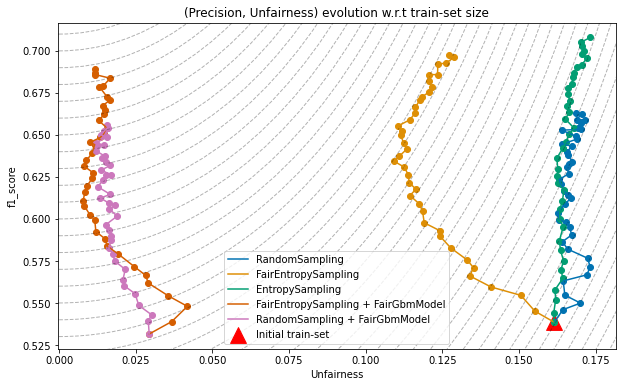}
\caption{($F_1$-score, Unfairness) evolution w.r.t. the training-set. Iterations start at the bottom-right corner of the graph. Top-left corner gives the best trade-off. Due to the finite pool-set, all methodologies involving $h$ ultimately reach the same endpoint ($h^{(fair)}$ remains completely fair).}
\label{fig:FALperformance_unfairness_}
\end{center}
\end{figure}

\paragraph*{Fair active learning with fairer model $h^{(fair)}$} 
We study the efficiency of our optimal fair classifier $h^{(fair)}$ (called \texttt{FairGbmModel} in the experiments) in terms of predictive performance and unfairness. Figure~\ref{fig:FALperformance_unfairness_} shows that our methodology \texttt{RandomSampling + FairGbmModel} and \texttt{FairEntropySampling + FairGbmModel} exhibits exact fairness (i.e. $\hat{\mathcal{U}}(h^{(fair)})\approx 0$) with slightly lower predictive performance compared to the base model. 
Additionally, in Figure~\ref{fig:FAL_FML_analysis}, we included the ex-post fairness calibration in AL scenarios as a baseline, employing \texttt{EntropySampling} (Left) and \texttt{DensityWeightedSampling} (Right). It is important to note that the ex-post results in each AL iteration do not influence the sampling strategies. Within our framework, ex-post fairness calibration is specifically designed for ML models trained on i.i.d. samples from $\mathcal{X}$ and is not optimized for models trained on sets selected via active sampling strategies. As expected, its performance declines beyond a certain AL iteration, as indicated by the opaque green curve in the figure. This baseline solution leads to the lowest fairness and accuracy given the labeling budget (see $\square$ in the figure). On the other hand, ex-post fairness calibration is inherent in PL (i.i.d. samples from $\mathcal{X}$) and is explicitly defined within the FAL framework, as these FAL techniques assume the fairness of the model before sampling informative instances.
All in all, our fair active learning framework achieves superior performance-unfairness trade-off.

In summary, Figure~\ref{fig:FALperformance_unfairness_} and Figure~\ref{fig:FAL_FML_analysis} show that FAL is a method that 1) collects both informative and fair examples and 2) proposes a well-balanced fair classifier with superior performance-unfairness compared to PL. As mentioned above, since all approaches with the base model $h$ result in the same trade-off between performance and unfairness once the entire pool-set is exhausted, FAL depletes fair instances after a certain point, leading to a loss of fairness. Therefore, we recommend stopping the process when there is no further progress in reducing unfairness.

\section{Conclusion}
The adoption of AI technology in the insurance industry has become essential for companies to remain competitive and meet evolving customer needs. However, achieving successful adoption requires addressing two critical objectives - the need for high-quality data and efficient learning models, as well as good governance to ensure fairness and proper interpretation of results. Numerical analysis and real-world examples have shown that active learning is an effective method to significantly decrease the amount of labeled data required to train an effective machine learning model. Nevertheless, naive active learning techniques may result in biased or unfair instances being selected, which could pose significant operational and reputational risks to insurance companies. By incorporating model-agnostic fair active learning, insurance companies can 1) collect informative and fair instances, and 2) propose a fairer model that offers exact fairness (ensuring that it does not discriminate or produce biased results) while maintaining competitive predictive performance. As the multi-class classification domain continues to expand, integrating active learning and fairness is becoming increasingly important. By adopting this approach, companies can benefit from enhanced AI performance, improved risk segmentation, improved decision making and increased operational efficiency, all while mitigating potential risks associated with biased or DP-unfair results. However, we note that some insurance applications may require tailored debiasing methods based on other ethical considerations, such as actuarial fairness as mentioned in the introduction. Since our approach to DP-fairness mitigation is query-agnostic, the framework presented in this study can serve as an initial step toward incorporating more specific unbiased constraints as needed.

A future direction of research is to investigate the use of Explainable AI (XAI) techniques in the active learning framework. XAI can provide insight into how machine learning models arrive at decisions and help identify potential sources of bias or discrimination. By interpreting the results of an active learning algorithm, XAI can build trust in the process and ensure decisions are made based on relevant and non-discriminatory factors. In addition, developing AI systems that can learn continuously from new data sources is another area of research interest. While active learning in an offline scenario is effective, it relies on a fixed set of training data, whereas the online learning can help AI systems adapt to evolving data sources in the insurance industry. However, online learning raises new challenges, such as how to maintain fairness (and transparency) as the model learns from new data sources.


\section*{Acknowledgments}
Our sincere thanks to Société Générale Insurance for their invaluable contribution to this project, including the provision of the (anonymized) NPS dataset. This R\&D initiative aims to enhance AI governance within Société Générale Insurance by promoting affordability and fairness in AI systems, in light of the growing democratization of ML model deployment in insurance. 
Furthermore, we sincerely thank the referees for their valuable feedback, especially for providing valuable references and offering insightful suggestions, notably, in the conduct of additional empirical studies. We believe these enhancements significantly improve the clarity and quality of the paper.

\bibliographystyle{alpha}
\bibliography{biblio}

\newpage
\appendix
\addcontentsline{toc}{section}{Appendices} 
\renewcommand{\thesubsection}{\Alph{subsection}}
\renewcommand{\thesubsubsection}{\Alph{subsection}.\arabic{subsubsection}}


\section*{Appendix}

This Appendix is dedicated to additional numerical experiments. In Section~\ref{appsec:addfigAL}, we present the effectiveness of traditional AL methods on the dataset NPS of Société Générale Insurance. Section~\ref{appsec:add_garage} is dedicated to the additional numerical results on the actuarial dataset \texttt{freMPL}. In particular, in section \ref{appsec:garage}, we justify why  the variable \texttt{garage} is an important factor in determining the auto pricing, while section \ref{appsec:add_garage_FALvsPL} extends some additional experiments from section \ref{subsec:FALresults}.

\subsection{Numerical experiments on the dataset NPS of Société Générale Insurance}
\label{appsec:addfigAL}

All the graphs presented in this section are averaged over 15 simulations and colored area corresponds to the standard deviation. Each simulation corresponds to an initial random training-set.

\paragraph{Dataset.} The textual data that we study is the NPS (Net Promoter Score) verbatims of Société Générale Insurance. The NPS is an indicator that measures the level of customer satisfaction. This indicator is measured through a survey carried out at regular intervals with a sample of customers. The survey is sent immediately to the customer following a call with a customer relations centers. The question asked to the customer to measure the level of satisfaction is the following:

"\textit{Would you recommend our insurance company to a friend or your family?}"

The answer to this question is a score ranging from 1 (detractor) to 10 (promoter). These scores constitute our label space $\{1, \cdots, K\}$ with $K$ a value between 2 (binary classification) and 10 (multi-class classification) depending on the split. We decided to study a binary task $\mathcal{Y} = \{0, 1\} = \{\rm{score} \leq c, \rm{score} > c\}$ with $c\in [K-1]$ being the \textit{imbalanced} parameter. We can choose $c$ such that we can range from a 10\% (\textit{imbalanced} case) to 50\% (\textit{balanced} case) imbalanced rate. In addition to the question the following question is also asked on surveys:

"\textit{Why did you give this rating?}"

The answer to this question is a free text comment allowing clients to justify the rating given by explaining the main reasons for their satisfaction or dissatisfaction. Note that the active learning intervention in the NPS dataset is only designed to assess the performance of the active learning approach for textual data. In this scenario, we already have access to the responses. When comparing active learning to the passive counterpart, we conceal most responses and selectively reveal some, following active learning guidelines. That being said, a construction of a numerical representation model of verbatims in a \textit{semantic space} allowing to measure the similarity of the verbatims has been implemented: for this purpose the algorithm \textit{Doc2vec} is used \cite{pmlr-v32-le14}. Unlike embedding algorithms of type \textit{Word2vec}  \cite{Mikolov13}, \textit{Doc2vec} has the advantage of providing in a "native" way a digital representation of an entire textual document (here a verbatim for example) without having to resort to an aggregation of the embedding of each of the words composing the verbatim. Each numerical representation of the textual documents represents our sample from $\mathcal{X}$. The sample made available for the active learning study consists of around $5000$ verbatims that correspond to a collection period of approximately one year (late April 2018 to late May 2019).

\begin{figure}
\begin{center}

\begin{subfigure}{\textwidth}
 \centering
\includegraphics[scale=0.68]{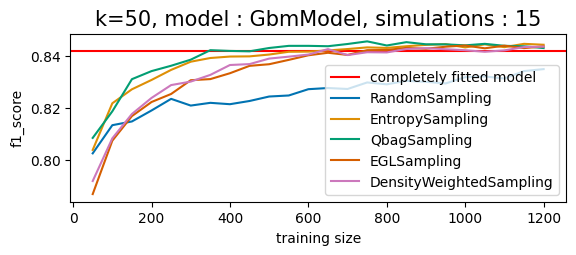}
\caption{}
\label{fig:al_plot_nps30_a}
\end{subfigure}

\begin{subfigure}{\textwidth}
 \centering
\includegraphics[scale=0.65]{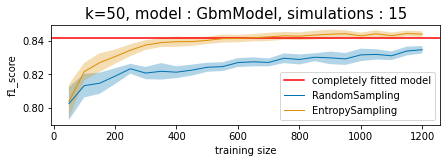}
\caption{}
\label{fig:al_plot_nps30_b}
\end{subfigure}

\caption{Performance of PL and AL on a balanced dataset.}
\label{fig:al_plot_nps30}
\end{center}
Performance of the classification procedures in terms of \textbf{$F_1$-score} for the \textbf{XGBoost} estimator w.r.t the training-set iteratively constructed by PL and AL methods mentioned above. In each iteration we query $k=50$ instances on a \textbf{balanced} dataset (30\%). Each line corresponds to the mean over 15 simulations and the colored area represents the standard deviation (Figure (b)).
\end{figure}

\paragraph*{Model performance.} Numerical studies in Figures \ref{fig:al_plot_nps30_a} and \ref{fig:al_plot_nps30_b} show that AL strategies outperform PL by sampling better quality data for the studied machine learning model (i.e. XGBoost). Indeed, most of the AL strategies converge to 0.84 with only 400 labeled data instead of 2000 with PL: in terms of model performance, AL is $5$ times more efficient than PL.

\begin{figure}[H]
\begin{center}
\begin{subfigure}{.55\textwidth}
  \centering
  \includegraphics[scale=0.45]{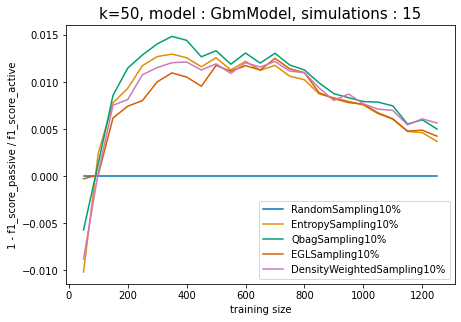}
    \caption{}
    \label{fig:imbalanced_perf_a}
\end{subfigure}%
\begin{subfigure}{.55\textwidth}
  \centering
  \includegraphics[scale=0.45]{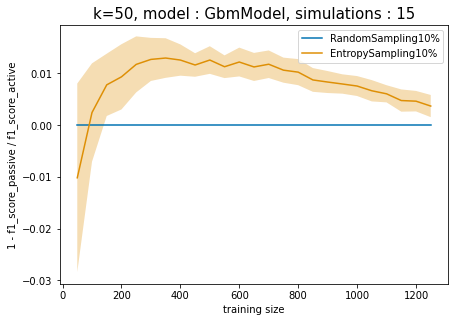}
    \caption{}
    \label{fig:imbalanced_perf_b}
\end{subfigure}
\caption{Performance of PL and AL methods on a imbalanced dataset (10\%).}
\label{fig:imbalanced_perf}
\end{center}
Performance of the classification procedures in terms of \textbf{$F_1$-score} for the \textbf{XGBoost} estimator w.r.t the training-set iteratively set by PL and AL methods on a \textbf{imbalanced} dataset (10\%) where for each iteration we query $k=50$ instances. Each line corresponds to the mean over 15 simulations and the colored area represents the standard deviation (Figure (b)).
\end{figure}

\paragraph*{Model robustness to imbalanced dataset.}
Figure \ref{fig:imbalanced_perf} highlights it by evaluating the gap in performance between AL and PL:

\begin{equation*}
    \text{GAP} = 1 - \dfrac{\text{f1\_score\_passive}}{\text{f1\_score\_active}}\enspace,
\end{equation*}

where $\text{f1\_score\_passive}$ (resp. $\text{f1\_score\_active}$) is the $F_1$-score of the passive learning (resp. active learning) process. The gap in performance between AL and PL shown in Figure \ref{fig:imbalanced_perf} indicates the efficiency of AL. Furthermore, we note that, the more unbalanced the data, the closer the performance of AL is to the performance of PL. Precisely \cite{Ertekin07ImbAct} shows that AL performs well to a slightly or moderately imbalanced case but can be inefficient to a heavily imbalanced one.



\subsection{Actuarial datatset \texttt{freMPL}: additional experiments}
\label{appsec:add_garage}

\subsubsection{Impact of garage on claims}
\label{appsec:garage}

\begin{figure}[H]
\begin{center}
\includegraphics[scale=0.6]{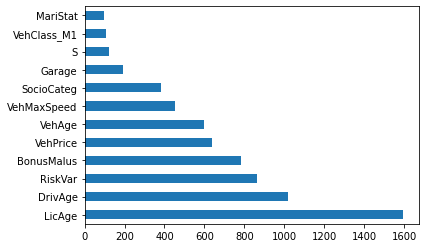}
\caption{Feature importance for claim prediction with XGBoost on \texttt{freMPL} dataset.}
\label{fig:feature_importance}
\end{center}
\end{figure}

\begin{figure}[H]
\begin{center}
\includegraphics[scale=0.65]{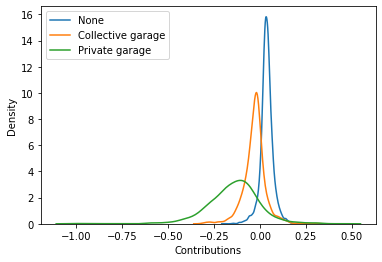}
\caption{Distribution of Shapley values with XGBoost on \texttt{freMPL} dataset given `garage = None', `garage = Collective' and `garage = Private'.}
\label{fig:contrib}
\end{center}
\end{figure}

Our study in this appendix focuses on the \texttt{freMPL} dataset, using XGBoost to predict claims. The model performs very well on both the test-set ($F_1$-score of $0.83$) and the train-set ($F_1$-score of $0.84$). We identify that the feature `garage` has a significant impact on claim prediction through our feature importance analysis in Figure~\ref{fig:feature_importance} (Among 27 features, both gender (`S`) and `garage` are ranked among the top 12 most important features)
, while Shapley values in Figure~\ref{fig:contrib} demonstrate the influence of garage distribution. Notably, we discover that private garages (`garage = Private`) have lower risks for claim prediction, suggesting the need for insurance companies to consider garage type when evaluating risks and premiums. Note that `garage = None` indicates on average higher risks for predicting claims, with a lower uncertainty in the prediction (greater mean and lower variance).

\paragraph*{About the Shapley values} Shapley values~\citep{lundberg2017unified} measure a feature's marginal contribution to the model output by comparing the output with and without the feature. They are calculated using a weighted average of all possible feature combinations and help explain how different features contribute to accurate predictions. See~\cite{shapley1953value, lundberg2017unified, sundararajan2020many} for more details. Note that Shapley values are one of the most commonly used methods of \textit{model explanation}.


\subsubsection{Fair active learning}
\label{appsec:add_garage_FALvsPL}

\paragraph{Fair active learning vs Passive learning}
As discussed in Section~\ref{subsec:FALresults} and also illustrated below in Figure~\ref{fig:fairALperfunfApp} and~\ref{fig:FALperformance_unfairnessApp}, FAL achieves a good balance between model performance and unfairness, outperforming PL on both fronts. However, \texttt{FairEGLSampling} stands out as an exception, showing lower efficiency than other FAL methods, possibly because it depends on the training loss, making it less model-agnostic. 

\begin{figure}[H]
\begin{center}
\begin{subfigure}{\textwidth}
 \centering
\includegraphics[scale=0.6]{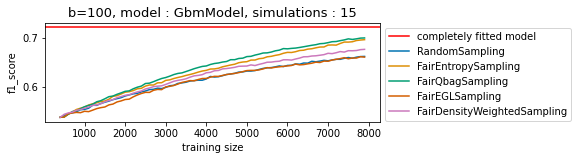}
\caption{}
\label{fig:fairALperfApp}
\end{subfigure}
\begin{subfigure}{\textwidth}
 \centering
\includegraphics[scale=0.6]{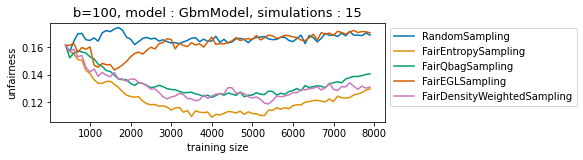}
\caption{}
\label{fig:fairALunfApp}
\end{subfigure}
\caption{Performance and unfairness of PL and FAL approaches on \texttt{freMPL} dataset.}
\label{fig:fairALperfunfApp}
\end{center}
Model evaluation ($F_1$-score and Unfairness) of PL and FAL for the model \textbf{XGBoost} w.r.t the training-set iteratively constructed with the sampling stategies mentionned above. At each iteration, we query $b=100$ instances.
\end{figure}

\begin{figure}[H]
\begin{center}
\includegraphics[scale=0.52]{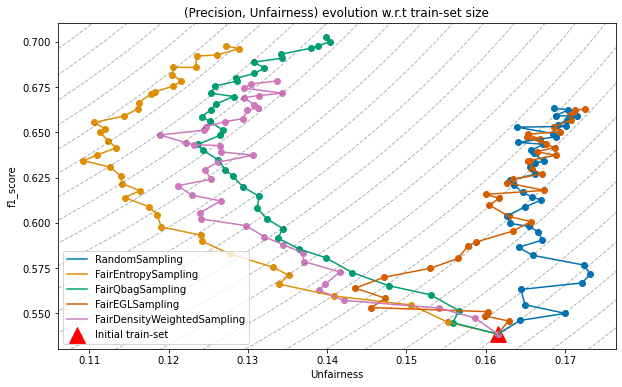}
\caption{($F_1$-score, Unfairness) evolution w.r.t. the training-set. Iterations start at the bottom-right corner of the graph. Top-left corner gives the best trade-off. \textbf{Remark:} Due to the finite pool-set, all methodologies ultimately reach the same endpoint.}
\label{fig:FALperformance_unfairnessApp}
\end{center}
\end{figure}

\paragraph{Fair active learning vs Active learning vs Passive learning}
\label{appsec:add_garage_FALvsALvsPL}

\begin{figure}[H]
\begin{center}
\includegraphics[scale=0.55]{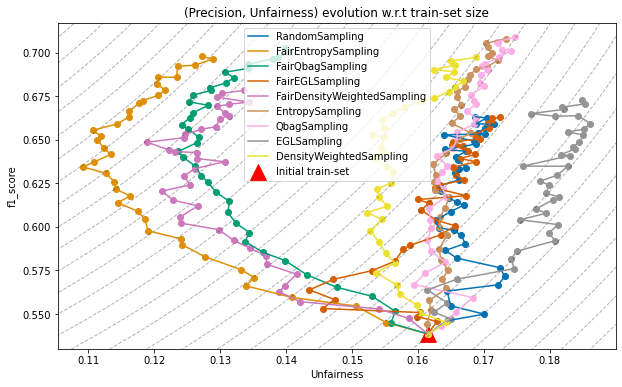}
\caption{($F_1$-score, Unfairness) evolution w.r.t. the training-set. Iterations start at the bottom-right corner of the graph. Top-left corner gives the best trade-off. \textbf{Remark:} Due to the finite pool-set, all methodologies ultimately reach the same endpoint.}
\label{fig:FALperformance_unfairnessApp1}
\end{center}
\end{figure}

Figure~\ref{fig:FALperformance_unfairnessApp1} shows that FAL offers the best compromise between model performance and fairness.

\appendix

\end{document}